%% file: main.tex
\documentclass[sigconf]{acmart}

\input{math_commands.tex}

\usepackage{url}
\usepackage{graphicx}
\newtheorem{definition}{Definition} 
\usepackage{enumitem}
\usepackage{minted}
\usepackage{xcolor} 
\usepackage{caption}    
\usepackage{graphicx}
\usepackage{amsmath}
\usepackage{multirow}
\usepackage{algorithm}
\usepackage{algpseudocode}  
\usepackage{xcolor}         
\usepackage{amsfonts}       
\usepackage{hyperref}       
\usepackage{circledsteps}
\usepackage{graphicx}
\usepackage{subfig}
\usepackage{caption}
\usepackage{enumitem}
\usepackage{tcolorbox}
\tcbuselibrary{breakable}
\usepackage{newfloat}
\DeclareFloatingEnvironment[name=Equation]{bnf}

\usepackage{makecell}

\usepackage{listings}
\lstdefinestyle{nettxt}{
  basicstyle=\ttfamily\scriptsize,
  breaklines=true,
  columns=fullflexible,
  frame=single
}

\input{macro}

\AtBeginDocument{%
  }

\setcopyright{acmlicensed}
\copyrightyear{2018}
\acmYear{2018}
\acmDOI{XXXXXXX.XXXXXXX}
\acmConference[Conference acronym 'XX]{Make sure to enter the correct
  conference title from your rights confirmation email}{June 03--05,
  2018}{Woodstock, NY}
\acmISBN{978-1-4503-XXXX-X/2018/06}

\begin{document}

\title{Understanding Network Behaviors through Natural Language Question-Answering}

\author{Mingzhe Xing, Chang Tian, Jianan Zhang, Lichen Pan, Peipei Liu, Zhaoteng Yan, Yinliang Yue}
\renewcommand{\shortauthors}{Trovato et al.}

\input{sections/00-abstract}

\maketitle

\input{sections/01-intro}

\input{sections/02-relatedwork}

\input{sections/03-prelmn}

\input{sections/04-method}

\input{sections/05-exp}

\input{sections/06-conclusion}

\bibliographystyle{ACM-Reference-Format}
\bibliography{ref}

\appendix

\input{sections/07-appendix}

\end{document}

%% file: math_commands.tex
\usepackage{amsmath,amsfonts,bm}

\def\eqref#1{equation~\ref{#1}}

\def\1{\bm{1}}

\DeclareMathAlphabet{\mathsfit}{\encodingdefault}{\sfdefault}{m}{sl}
\SetMathAlphabet{\mathsfit}{bold}{\encodingdefault}{\sfdefault}{bx}{n}



%% file: macro.tex
\usepackage{xspace}
\newcommand{\eg}{\textit{e.g.,}\xspace}
\newcommand{\ie}{\textit{i.e.,}\xspace}
\newcommand{\ours}{\texttt{NetMind}\xspace}

%% file: sections/00-abstract.tex
\begin{abstract}
Modern large-scale networks introduce significant complexity in understanding network behaviors, increasing the risk of misconfiguration.
Prior work proposed to understand network behaviors by mining network configurations, typically relying on domain-specific languages interfaced with formal models.
While effective, they suffer from a steep learning curve and limited flexibility.
In contrast, natural language (NL) offers a more accessible and interpretable interface, motivating recent research on NL-guided network behavior understanding.
Recent advances in large language models (LLMs) further enhance this direction, leveraging their extensive prior knowledge of network concepts and strong reasoning capabilities.
However, three key challenges remain: 
1) numerous router devices with lengthy configuration files challenge LLM's long-context understanding ability;
2) heterogeneity across devices and protocols impedes scalability;
and 3) complex network topologies and protocols demand advanced reasoning abilities beyond the current capabilities of LLMs.
To tackle the above challenges, we propose \ours, a novel framework for querying networks using NL.
Our approach introduces a tree-based configuration chunking strategy to preserve semantic coherence while enabling efficient partitioning.
We then construct a unified fact graph as an intermediate representation to normalize vendor-specific configurations.
Finally, we design a hybrid imperative-declarative language to reduce the reasoning burden on LLMs and enhance precision.
We contribute a benchmark consisting of NL question-answer pairs paired with network configurations. 
Experiments demonstrate that \ours achieves accurate and scalable network behavior understanding, outperforming existing baselines.
\end{abstract}

%% file: sections/01-intro.tex
\section{Introduction}
Deep learning has facilitated various real-life scenarios~\cite{tian2022anti,li2021paint4poem,tian2024fighting,jelaca2025automated,schildermans2025structured,zhou2025controllable}, including modern computer networks.
The scale and complexity of modern computer networks continue to grow exponentially, driven by increasingly diverse web functionalities and applications~\cite{xing2024understanding,tian2025using,tian2025large,tian2024generic}.
For instance, Amazon Web Services operates global-scale data centers across 120 availability zones, supporting massive router deployments to sustain global connectivity and performance\footnote{https://aws.amazon.com/about-aws/global-infrastructure/}. 
This scale significantly complicates the understanding of network behavior, thereby increasing the risk of misconfigurations that can lead to large-scale service disruptions.
A notable example is the 2021 Facebook outage\footnote{https://engineering.fb.com/2021/10/04/networking-traffic/outage/}, where a misconfigured Border Gateway Protocol~(BGP)~\cite{rekhter2006rfc} route announcement rendered the service inaccessible across the global Internet, underscoring the critical need for correctly understanding network behaviors.

Prior work~\cite{birkner2020config2spec,kheradmand2020automatic} analyzed network behaviors through formal modeling of network configurations.
These models are typically interfaced with \textit{Domain Specific Language~(DSL)}, which are machine-readable and suitable for formal reasoning.
However, DSLs suffer from poor scalability and a steep learning curve~\cite{han2025network}.
To address this, researchers turn to use \textit{Natural Language~(NL)} to interact with network, which provide greater flexibility and \textit{human-readable insights for human operator}.
Net2Text~\cite{birkner2018net2text} parses NL questions with predefined rules and generates summaries of relevant network traffic behaviors.
Lumi~\cite{jacobs2021hey} extracts entities from NL instructions and synthesizes corresponding network configurations. 
These efforts highlight the potential of NL-based approaches in improving the accessibility and interpretability of network behavior analysis.



Recent advances in Large Language Models (LLMs) \cite{zhao2023survey} have shown their superior capability in text understanding and reasoning, prompting their application to a range of networking tasks, \eg configuration translation and synthesis~\cite{mondal2023llms,ding2024poster,wu2024netllm,wang2024netconfeval,mekrache2024intentbased,fang2024large,kou2025gia}.
However, the potential of using LLM to support network behavior understanding remains largely underexplored.
In this paper, we identify three key challenges must be addressed to fully realize this potential.
\textbf{Challenge 1. Understanding and reasoning over long-context configurations in large-scale networks.}
Real-world networks typically comprise tens to hundreds of routers, each configured with thousands lines of functions. 
This imposes understanding and reasoning burdens on LLMs, or even exceeding their context window limits~(the configuration files of the smallest network in our experiment exceed 20k tokens). 
\textbf{Challenge 2. Heterogeneity of network devices and functionalities limits scalability.}
Routers from multiple vendors (\eg Cisco, Huawei, and Juniper) introduces substantial variability in configuration formats~\cite{martinez2014network}.
Additionally, network functions span diverse configuration primitives, including definitions of interface, IP prefix lists, route policies, and protocol configurations such as Open Shortest Path First (OSPF)~\cite{moy1998ospf} and BGP etc, which poses additional normalization challenges.
\textbf{Challenge 3. Complex topologies and protocol interactions demand advanced reasoning capabilities.}
Directly answering questions from raw configuration files~\cite{wei2025intaintentbasedtranslationnetwork} is infeasible in large-scale networks with intricate routing policies and interdependent protocols.
Alternatively, generating code for configuration parsing and reasoning may solve this issue, but prior work~\cite{wang2024netconfeval} shows that ensuring semantic and functional correctness in such LLM-generated code remains an open challenge.

To overcome the aforementioned challenges, we propose \ours, a novel LLM-driven framework for understanding network configurations and answering NL questions.
Our approach consists of three key innovations:
First, we introduce a \textbf{tree-based chunking strategy} to efficiently process numerous and lengthy configuration files.
By identifying semantically coherent configuration blocks and linking them via reference-based dependencies (\eg route policy or IP prefix list invocations), we construct a configuration syntax tree. 
Traversing root-to-leaf paths in this tree yields semantically self-contained segments, enabling efficient and context-preserving processing by LLMs.
Second, we design a \textbf{unified, vendor-agnostic fact graph intermediate representation} to extract fine-grained factual knowledge from configurations and construct the topological structure of networks as graphs.
Specifically, LLMs are employed to {translate heterogeneous configuration files into explicit facts}, from which implicit facts are further deduced based on existing facts and protocol rules.
These facts are then converted to a fact graph intermediate representation, where the nodes and edges denotes the facts and their relationships, respectively.
This representation serves as a unified foundation for subsequent reasoning, supporting both symbolic knowledge inference and graph-based computation.
Third, instead of directly answering the questions or generating solution code, we design \textbf{a hybrid imperative-declarative query language} that allows LLMs to express reasoning processes more effectively.
This language disentangles and offloads routine yet complicated computing functions to the declarative part while using imperative language to perform more flexible reasoning.
LLMs generate programs in this hybrid language, which are then executed within a dedicated runtime environment to produce the results.

To summarize, our contributions are as follows: 
\begin{itemize}[leftmargin=*, noitemsep]
    \item To the best of our knowledge, this work presents the first NL interface for querying network configurations and reasoning about network behaviors. Furthermore, we develop a novel benchmark that pairs router configurations with expert-verified NL question-answer pairs, enabling systematic evaluation of network understanding systems.
    \item We first design a tree-based, vendor-agnostic configuration chunking strategy that preserves syntactic and semantic coherence while reducing input length. We propose to construct a fact graph as intermediate representation for unifying heterogeneous, vendor-specific configurations. Moreover, a novel hybrid imperative-declarative language is proposed to improve reasoning accuracy.
    \item Comprehensive experiments conducted on the proposed benchmark demonstrates that \ours significantly outperforms baselines and effectively supports network behavior understanding.
\end{itemize}

%% file: sections/02-relatedwork.tex
\section{Related Work}
\subsection{Network Configuration Modeling}

Efforts for modeling network configurations can be broadly divided into three categories, \ie formal-method-based, graph-based and AI-based methods.
VeriCon~\cite{ball2014vericon} uses first-order logic to formally specify network topologies and desired properties, and employs deductive verification techniques to verify network status. 
Batfish~\cite{fogel2015general} and Mineswepper\cite{beckett2017general} employ Datalog~\cite{ceri1989you} and Binary Decision Tree~\cite{laurent1976constructing} to formally model routing polices, and utilize Satisfiability Modulo Theories~\cite{barrett2018satisfiability} solver to check if networks satisfy given properties.
Beyond classical graph traversal algorithms such as Breadth First Search and Depth First Search algorithms on topology graphs to learn routes, ARC~\cite{gember2016fast} proposes a high-level abstraction in the form of weighted digraphs for network configurations, and identifies network properties through advanced graph algorithms.
Tiramisu~\cite{abhashkumar2020tiramisu} models adjacency and forwarding graphs, and deduces network connectivity by calculating the adjacency and paths of graphs.
Lightyear~\cite{tang2023lightyear} constructs BGP topology graphs, and adopts a modular approach to perform configuration checking.
CEGS~\cite{liu2025cegs} uses AI-based method, \ie Graph Neural Network~\cite{wu2020comprehensive}, to generalize from pre-designed examples to arbitrary topology, and an LLM to synthesize configuration.
Network CoPilot~\cite{han2025network} exploits NL-driven intent to specify network requirements, and generates or updates network configurations using LLMs.

While AI- and LLM-based methods have shown superior accuracy in network modeling, they still suffer from the long-context understanding and reasoning, heterogeneous configurations, and complex topological reasoning issues.

\subsection{Natural Language in Networking}
Natural language (NL) interaction with networks offers a promising approach for improving operator efficiency, due to its interpretability and human-readability.
\citet{alsudais2017hey} collects typical network tasks and applies Part-of-Speech tagging~\cite{martinez2012part} to impose syntactic structure on operator-issued NL inputs.
Net2Text~\cite{birkner2018net2text} designs a set of rules to parse NL questions, enabling network-wide reasoning over packet forwarding behaviors.
The emergence of LLMs have spurred exploration of their potential for networking tasks.
COSYNTH~\cite{mondal2023what} employs LLMs to synthesize network configurations and find that the generated outputs often contain semantic and syntactic errors, highlighting the importance of formal verification.
NetConfEval~\cite{wang2024netconfeval} builds a benchmark to evaluate the capabilities of LLMs in intent translation and code generation, revealing significant gaps in generating valid networking code even with state-of-the-art models.
NetLLM~\cite{wu2024netllm} introduces various networking heads at the output side of LLM to generate answers for solving specific prediction or scheduling networking tasks.

Despite these efforts, network behavior understanding remains largely underexplored.
Moreover, existing LLM-driven approaches primarily focus on empirical studies of small-scale networks, neglecting the {heterogeneity} and {long-context dependencies} issues in real-world configurations.


%% file: sections/03-prelmn.tex
\section{Preliminary}


\subsection{Network Protocols and Configurations}
Each router within a network runs routing protocols, such as Open Shortest Path First (OSPF)~\cite{moy1998ospf} and Border Gateway Protocol (BGP)~\cite{rekhter2006rfc}, to exchange reachability information with neighbors and derive local forwarding rules.
This behavior is governed by configuration files, typically written in vendor-specific languages (\eg Cisco IOS) by network operators. 
As illustrated in Fig.~\ref{fig:bgp_example}, configurations are organized into logical blocks associated with specific protocols, interfaces, or routing policies.

OSPF is a link-state interior gateway protocol that computes shortest paths within an autonomous system (AS). 
Routers exchange link-state advertisements (LSAs) that describe local connectivity and link attributes such as bandwidth or administrative cost. 
These LSAs are flooded throughout the AS, allowing every router to construct a consistent view of the network topology.
Using this graph, each router independently compute shortest paths, where link weights are determined by configurable OSPF cost parameters.

In contrast, BGP functions as a path-vector protocol designed for inter-AS routing. 
It enables ASes to announce reachability to IP prefixes and negotiate routing policies with peers. 
BGP speakers exchange route announcements containing attributes such as local preference, AS path length, origin type, multi-exit discriminator, and interior gateway protocol cost.
Upon receiving routes, a BGP router applies import policies that may filter, modify, or prioritize routes before selecting the optimal path according to a standardized decision process based on these attributes. 
The selected route is then subject to export policies before being propagated to neighboring ASes. 
This mechanism supports flexible policy-based routing while maintaining global reachability.







\begin{figure}[tbp]
\centering
\begin{minipage}{0.49\columnwidth}
\begin{lstlisting}[style=nettxt]
hostname R1

interface lo0
 ip address 1.1.1.11 255.255.255.255

interface GigabitEthernet1
 ip address 1.0.25.2 255.255.255.0
 ip ospf cost 19

router ospf 1
 network 1.1.1.11 0.0.0.0 area 0
 network 1.0.25.2 0.0.0.0 area 0
\end{lstlisting}
\end{minipage}
\hfill
\begin{minipage}{0.49\columnwidth}
\begin{lstlisting}[style=nettxt]
hostname R2

route-map rm1 permit 10
 match ip address 1.1.1.18
 set local-preference 2
 set as-path prepend 1
 set metric 1

router bgp 1
 neighbor 1.1.1.5 remote-as 1
 neighbor 1.1.1.5 advertisement-interval 0
 neighbor 1.1.1.5 route-map rm1 out
\end{lstlisting}
\end{minipage}
\caption{OSPF and BGP configuration examples.}
\label{fig:bgp_example}
\end{figure}

\subsection{Imperative and Declarative Languages}
\label{sec:datalog}
Imperative programming specifies \textit{how a computation should be performed} by explicitly defining a sequence of instructions. 
Its execution proceeds step-by-step, with control flow and mutable state playing central roles. 
This paradigm requires detailed procedural reasoning about program behavior. 
Widely adopted languages such as C++, Java, and Python are primarily imperative, often incorporating object-oriented constructs to manage complexity.

In contrast, declarative programming abstracts \textit{what the desired outcome should be}, rather than how to achieve it. 
It emphasizes specification over control flow, thereby facilitating high-level reasoning about program logic and intent.
This abstraction enhances readability, modularity, and predictability, particularly in domain-specific applications. 
Declarative paradigms are prevalent in database querying (\eg SQL), user interface markup (\eg HTML, CSS) and etc.
Datalog~\cite{maier2018datalog} is a representative declarative logic programming language.
It provides a concise and expressive framework for specifying complex relationships and inference rules. 
Datalog programs consist of facts and rules, which say how to deduce new facts from known facts, thereby supporting a purely declarative reasoning process.
Owing to its simplicity, formal clarity, and strong theoretical guarantees, Datalog has found widespread use in deductive databases, program analysis, knowledge representation, and network configuration verification.

%% file: sections/04-method.tex
\begin{figure*}
    \centering
    \includegraphics[width=\textwidth]{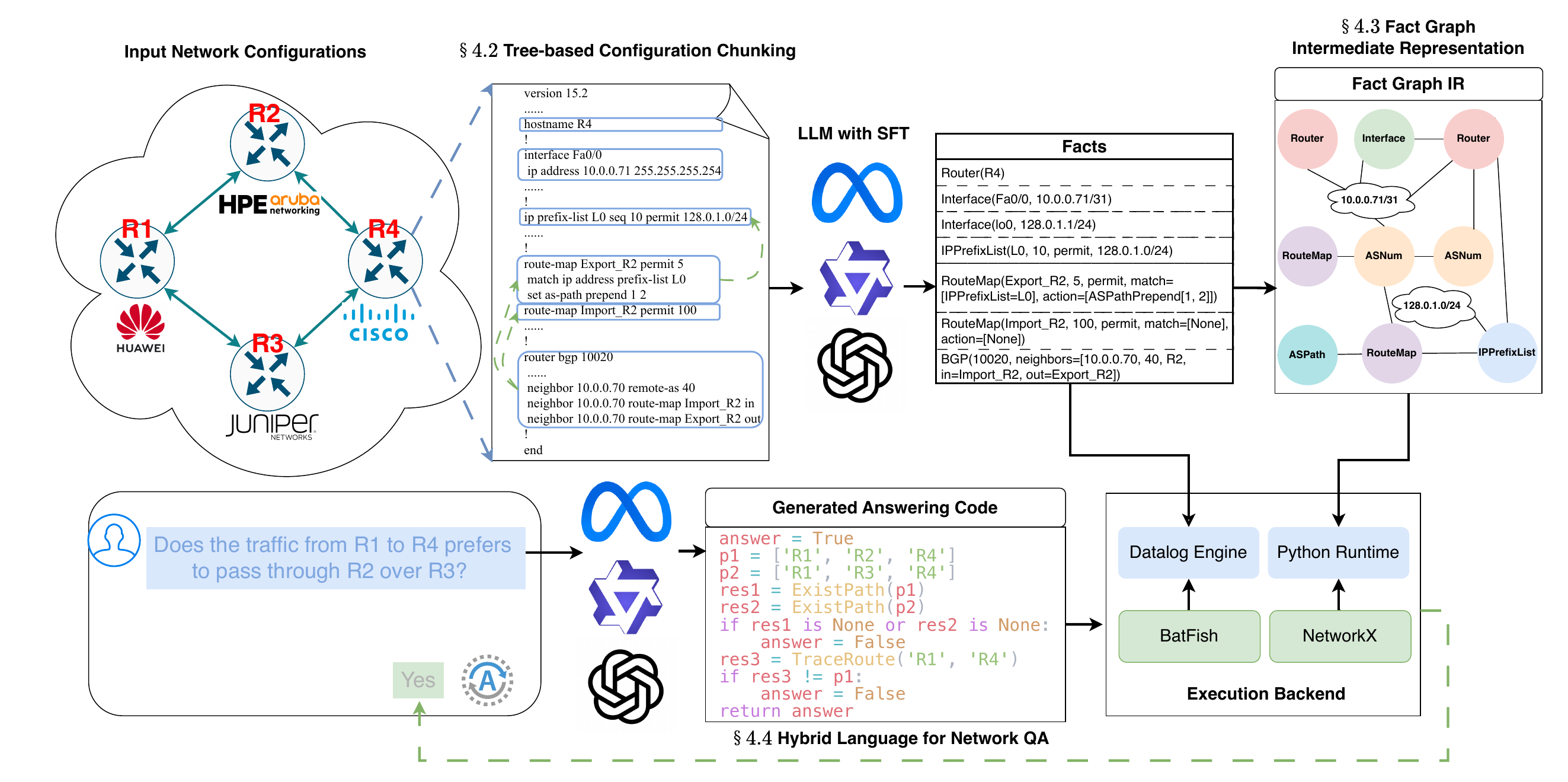}
    \caption{The overall workflow for answering NL questions about network behaviors. The tree-based chunking algorithm first decomposes lengthy configurations into semantic-coherent paths, which are then mapped to explicit facts. Implicit facts are deduced from rules to construct a fact graph IR. This IR supports to perform declarative and imperative reasoning over the generated hybrid code.}
    \label{fig:model}
\end{figure*}

\begin{algorithm}[tbp]
\caption{Tree-based Configuration Chunking}
\label{alg:tree_chunking}
\begin{algorithmic}[1]
\Require Configuration file $c$ in vendor-specific syntax
\Ensure Set of semantically coherent chunks $\mathcal{P}$

\Statex \textbf{Stage 1: Configuration Blocks Identification}
\State Split configuration $c$ by syntactic boundaries (\eg comments `\texttt{!}', indentation changes) to blocks $\mathcal{B} = \{b_1, b_2, \dots, b_n\}$
\For{each block $b_i \in \mathcal{B}$}
    \State Extract block identifier using regular expressions
\EndFor

\Statex \textbf{Stage 2: Dependencies Linking}
\State Initialize empty directed tree $T = (M, N)$
\State Let $M \gets \mathcal{B}$  \Comment{Each block is a node}
\For{each block $b_i \in \mathcal{B}$}
    \For{each identifier $id$ referenced in $b_i$}
        \For{each block $b_j \in \mathcal{B} \setminus \{b_i\}$}
            \If{$b_j$ defines $id$}
                \State Add edge $(b_i, b_j)$ to $N$  \Comment{$b_i$ depends on $b_j$}
            \EndIf
        \EndFor
    \EndFor
\EndFor

\Statex \textbf{Stage 3: Dependency Path Extraction}
\State Identify root nodes $\mathcal{R} \subseteq M$
\State Initialize $\mathcal{P} \gets \emptyset$
\For{each root $r \in \mathcal{R}$}
    \State Perform depth-first traversal from $r$ to all leaves
    \For{each root-to-leaf path $\pi = [r = m_1, m_2, \dots, m_k]$}
        \State Concatenate blocks in $\pi$ to form chunk $p$
        \State $\mathcal{P} \gets \mathcal{P} \cup \{p\}$
    \EndFor
\EndFor

\State \Return $\mathcal{P}$
\end{algorithmic}
\end{algorithm}
\section{Methodology}
In this section, we first give a formal definition of the network behavior understanding task.
After that, we introduce the overall framework of \ours as depicted in Fig.~\ref{fig:model}.
It consists of three modules, \ie tree-based chunking algorithm, fact graph intermediate representation transformation, and hybrid solution code generation for answering network questions.

\subsection{Problem Definition}
The task focused in this paper is to assistant network operators to easily understand network behavior through user-friendly natural language interaction, without manual inspection of low-level configuration details.
Formally, let the network be represented as a graph $G=⟨\mathcal{V},\mathcal{E}⟩$, where $\mathcal{V}$ denotes the set of routers and $\mathcal{E}\in \mathcal{V}\times \mathcal{V}$ represents the physical links between them.
Each router $v\in \mathcal{V}$ is associated with a configuration file $c_v\in \mathcal{C}$, where $\mathcal{C}$ is the set of router configuration files in the network snapshot. 
Given a NL question $Q$, the goal is to derive the answer $A=\ours (G, \mathcal{C}, Q)$.

\subsection{Tree-based Configuration Chunking}
\label{sec:chunk}
To preserve semantic integrity, we introduce a tree-based chunking algorithm, inspired by Abstract Syntax Tree-based chunking method used in code generation~\cite{gong2024ast}.
This algorithm decomposes network configurations into semantically coherent blocks while maintaining cross-block dependencies, enabling effective context management for LLMs. 
The approach proceeds in three stages:
\paragraph{\Circled{1} Configuration Blocks Identification.}
We begin by identifying fine-grained configuration blocks, which are syntactically and functionally distinct segments of configurations that encapsulate specific network functionalities (\eg interface definitions, routing policies).
As illustrated in the Cisco configuration snippet in Fig.~\ref{fig:model}, these blocks~(framed in blue) are typically delineated by structural cues such as indentation levels and comment delimiters (\eg the `\texttt{!}` token in Cisco syntax). 
To ensure robustness across heterogeneous, vendor-specific configuration formats, we employ vendor-agnostic regular expressions to accurately segment configuration files into functionally meaningful blocks.

\paragraph{\Circled{2} Dependencies Linking.}
Configuration blocks often exhibit interdependencies through symbolic references. 
For instance, as shown by the green dotted curves in Fig.~\ref{fig:model}, a BGP policy may invoke a route map (\eg \texttt{Export\_R2\_out}), which in turn references an IP prefix list (\eg \texttt{L\_0}). 
We recover these cross-block dependencies through reference term matching, \ie resolving and matching identifiers (\eg policy names, prefix list names) across block boundaries using regular expression-based pattern matching. 
This process constructs a directed dependency tree, where nodes correspond to configuration blocks and edges denote referencing relationships among them.

\paragraph{\Circled{3} Dependency Path Extraction.}
From the constructed dependency tree, we extract all root-to-leaf traversal paths, each forming a semantically self-contained  configuration sequence. 
Every path contains both the usage and full definition of all referenced entities, thereby preserving contextual completeness. 
These dependency paths serve as semantically enriched input chunks for downstream LLM processing, reducing context fragmentation and improving reasoning accuracy.
Notably, our method is lightweight and scalable, making it particularly suitable for real-world and multi-vendor environments. 
The pseudocode of the proposed algorithm can be found in Algorithm~\ref{alg:tree_chunking}.

\subsection{Fact Graph Intermediate Representation}
\label{sec:fgir}
In the previous part, we derive semantically coherent dependency paths through tree-based chunking algorithm.
In this section, we aim to transform these extracted paths from heterogeneous, vendor-specific configurations into a unified, vendor-agnostic intermediate representation~(IR) to facilitate downstream reasoning.
To this end, we propose the \textit{fact graph IR}, a structured knowledge representation that captures both explicit configuration statements and implicitly derived network behaviors.
The construction of the fact graph IR consists of three stages, \ie extracting explicit facts from raw configurations with LLMs, deducing implicit facts based on existing facts and protocol rules, and finally building the fact graph.

\begin{figure}[tbp]
\small
\[
\begin{aligned}
\langle program \rangle         & ::= \langle py\_stmt \rangle \\
                                & \quad \mid\ \langle datalog\_stmt \rangle \\
\langle py\_stmt \rangle        & ::= \text{any valid Python statement} \\
\langle datalog\_stmt \rangle   & ::= \texttt{query} ~ \{~ \langle predicate \rangle ~\} \\
\langle predicate \rangle & ::= \texttt{Reach}~(~\langle router \rangle,~\langle router \rangle~) \\
                                & \quad \mid\ \texttt{ExistPath}~(~\langle router\_list \rangle~) \\
                                & \quad \mid\ \cdots \\
                                & \quad \mid\ \texttt{AllPath}~(~\langle router \rangle,~\langle router \rangle~) \\
                                & \quad \mid\ \texttt{TraceRoute}~(~\langle router \rangle,~\langle router \rangle~) \\
                                & \quad \mid\ \texttt{ShortestOSPFPath}~(~\langle router \rangle, \\
                                & \qquad\quad \langle router \rangle~) \\
\langle router\_list \rangle    & ::= [~\langle router \rangle ~ \{~,~ \langle router \rangle ~\}~] \\
\langle router \rangle          & ::= \text{Identifier}~(\eg~\texttt{R\_1},~\texttt{R\_2})
\end{aligned}
\]
\caption{BNF syntax definition of the hybrid language.}
\label{eq:dsl_syntax}
\end{figure}

\paragraph{\Circled{1} Explicit Facts Extraction.} 
Considering that network configurations typically declare what the network routing behavior should be, we design a set of fact base to present these \textit{declared factual knowledge}.
It is inspired by Datalog language~\cite{maier2018datalog}~(introduced in \S\ref{sec:datalog}), which separates data (facts) knowledge from logic (rules) reasoning.
We model network configurations using a Datalog-like fact base and rule set.
\begin{definition}[Fact Base]
    $\mathcal{F}$ is a set of facts $f(a_1, \dots, a_n)$, where $f$ is a predicate denoting network identifiers, \eg \texttt{Router}, \texttt{IPPrefixList}, \texttt{RouteMap}, \texttt{OSPFNetwork}, \texttt{OSPFCost} etc, and $a_i$ is a constant argument (\eg router name, IP prefix). The complete schema of $\mathcal{F}$ is provided in Appendix~\ref{app:facts}.
\end{definition}
A fact example is \texttt{Router(\text{R1})}, which asserts the existence of a router named \text{R1}.
Configuration paths explicitly declare such facts, thus is suitable for leveraging LLM to exract facts from them, guided by a carefully designed prompt (see Appendix~\ref{app:prompt_template} for details).
To further enhance extraction accuracy, we employ low-rank adaptation (LoRA)~\cite{hu2022lora} to fine-tune foundation LLMs with configuration-to-fact pairs annotated by human.

\paragraph{\Circled{2} Implicit Facts Deduction.}
While \textit{explicit facts} capture declared configuration elements, many network behaviors (\eg routing adjacencies) emerge only through protocol logic. 
To infer such \textit{implicit facts}, we define a \textit{rule set} $\mathcal{H}$ grounded in network protocols.
\begin{definition}[Rule Set]
    $\mathcal{H}$ is a set of rules $h(a_1, \dots,a_n) \leftarrow b_1 \otimes b_2 \otimes \cdots \otimes b_n$, where the head expression $h$ is the derived fact, body expression $b$ checks for the presence of a fact or satisfaction of a condition, and $\otimes$ denotes logical operator including `$\land$`, `$\lor$`, and `$\lnot$` etc. These rules encode network protocols and can be used to deduce new facts based on known facts.
\end{definition}
For example, the existence of a routing edge between routers R1 and R2~(\ie the \texttt{RouteEdge} fact) can be deduced if they are connected via interfaces on the same subnet, which can be specified as the following rule:
\begin{align}
    \texttt{RouteEdge}(R_1, R_2) \leftarrow
 &\land\ \texttt{Interface}(R_1, \text{IP}_1/\text{Mask}), \nonumber\\
    & \land\ \texttt{Interface}(R_2, \text{IP}_2/\text{Mask}), \nonumber\\
    & \land \ 	\lnot\ R_1 == R_2, \nonumber\\
    & \land\ \texttt{Subnet}(\text{IP}_1/\text{Mask}) == \texttt{Subnet}(\text{IP}_2/\text{Mask}). \nonumber
\end{align}
By applying such rules to the extracted fact base $\mathcal{F}$, we derive new facts including \texttt{RouteEdge} and \texttt{BGPPeer}. 
The complete rule set is detailed in Appendix~\ref{app:facts}.

\paragraph{\Circled{3} Fact Graph IR Construction}
Drawing inspiration from graph-based network configuration synthesis work~\cite{beurer2022learning,han2024netren}, we construct a fact graph IR $G_F=(\mathcal{F}, \mathcal{R})$ based on the explicit and implicit facts, where the nodes $\mathcal{F}$ and edges $\mathcal{R}$ denote facts and their relations, respectively.
An example graph is shown in Fig.~\ref{fig:model}, which integrate facts including routers, interfaces, subnets, and policies into a unified topology.
As depicted, the \texttt{RouteEdge($R1$, $R2$)} connect nodes \texttt{Router($R1$)} and \texttt{Router($R2$)}, which in turn both connect to the same \texttt{SubNet(IP/Mask)} node.
This representation offers two key advantages:
1) it encodes the routing topology as a graph, enabling the application of graph algorithms (\eg shortest path finding) to answer complex network questions.
2) It functions simultaneously as a knowledge graph and a Datalog-compatible fact base, preserving both structural and logical semantics of the network, thus supporting accurate and interpretable symbloic reasoning.

\begin{figure}[tbp]
\begin{lstlisting}[language=Python]
[
frame=lines,
framesep=1.5mm,
baselinestretch=1.2,
breaklines=true,
breakanywhere=true,
linenos
]{python}
answer = True
p1 = ['R1', 'R2', 'R3']
p2 = ['R1', 'R4', 'R5', 'R3']
exist_p1 = ExistPath(p1) # Datalog Statement
exist_p2 = ExistPath(p2)
if exist_p1 is None or exist_p2 is None:
    answer = False
res = TraceRoute('R1', 'R3') # Datalog Statement
if res != p1:
    answer = False
return answer
\end{lstlisting}
\caption{An exemplified code for answering if the traffic from R1 to R3 prefers path $p1$ over $p2$.}
\label{code:order_example}
\end{figure}

\subsection{Hybrid Language for Network QA}
\label{sec:hybrid_code}
As LLMs exhibit strong understanding and reasoning abilities on networking tasks, a naive way for answering network questions is to directly feed the configurations or fact graph IR into LLM, and prompt it to generate end-to-end answers.
However, this is infeasible for real-world network, which typically comprise numerous routers and are governed by diverse, complex routing policies.
Therefore, it is necessary to transform NL questions to scalable solution code to reason the answer.
Net2Text collects network traffic data and employs a \textit{declarative code} (\ie SQL queries) to extract structured information. 
This approach is limited to predefined data fields and lacks the expressive power needed for sophisticated reasoning or computation.
In contrast, NetConEval adopts Python, a representative \textit{imperative code}, to encode network reasoning logic. 
While more expressive, the experiments indicate that generating syntactically and functionally correct network programs remains challenging even for state-of-the-art LLMs like GPT-4.

To address these limitations, we propose a hybrid language $\mathcal{L}=(\mathcal{L}_{decl}, \mathcal{L}_{imp}, \mathcal{I})$ that combines the strengths of both declarative and imperative paradigms.
It decouples query-oriented reasoning steps and offloads them to declarative parts $\mathcal{L}_{decl}$, while preserving the flexibility of imperative logic $\mathcal{L}_{imp}$ for complex reasoning tasks.
$\mathcal{I}$ serves as the execution backend to support both imperative and declarative execution engine.
Formally, we define the syntax of hybrid language as shown in Fig.~\ref{eq:dsl_syntax}. 
The Datalog predicates used for reasoning network questions can be found in Appendix\S\ref{app:bnf}.

Specifically, we implement two backends to execute the generated code.
The first employs a Datalog solver to deduce answers from existing facts and routing policies.
The second supports general logical computing in native Python, \eg numeric comparison, and simulates routing policies on the fact graph with graph algorithms, \eg Bellman-Ford~\cite{magzhan2013review} to calculate OSPF weights.
These well-specified routing policies are well-suited to algorithmic simulation, enabling accurate and verifiable reasoning.
As depicted in Fig.~\ref{fig:model}, the declarative and imperative components in the generated codes are computed by the two respective backend within a wrapped Python runtime $\mathcal{I}$ to produce the final answer.




%% file: sections/05-exp.tex
\section{Experiments}

\subsection{Benchmark Construction}
\paragraph{Network Topologies} We obtain network topologies from Topology Zoo~\cite{6027859}, a publicly available repository of real-world network. 
From this dataset, we randomly select 15 representative topologies and categorize them by scale: \textit{Small} (32–39 routers), \textit{Medium} (68–74 routers), and \textit{Large} (145–197 routers), with five topologies in each category. 
This stratification enables evaluation of system performance under increasing topological complexity.

\paragraph{Network Configuration Generation}
To generate semantically meaningful network configurations, we employ NetComplete~\cite{el2018netcomplete}, a network configuration synthesis tool. 
For each topology, we define a set of network requirements, including:
1) \textbf{ExistPath}: Ensures the existence of a valid route between two nodes;
2) \textbf{OrderedPath}: Specifies a preference ordering between two route paths (\eg path $P_1$ preferred over $P_2$ );
3) \textbf{LoadBalance}: Enforces equal preference for two route paths to distribute traffic evenly;
4) \textbf{KConnected}: Requires the existence of $K$ disjoint route paths between two nodes for redundancy.
Given these requirements presented in specified format, NetComplete synthesizes Cisco-style configuration files.
To assess robustness under varying requirement complexity, we vary the requirement number per network in $\{1,2,4,8,16\}$. 
To reflect real-world heterogeneity, we randomly transform the configurations into other major vendors' format: Cisco~(no change), Huawei, Juniper, H3C, and HPE Aruba. 
The transformation probabilities are set as $[0.4,0.3,0.1,0.1,0.1]$, reflecting approximate market shares~\footnote{https://my.idc.com/getdoc.jsp?containerId=prUS52590024}. 
These transformations are performed using Qwen3-32B, followed by cross-validation and correction by experienced network operators to ensure syntactic correctness and semantic fidelity.

\paragraph{QA Generation}
Building upon the network requirements and synthesized configurations, we construct a natural language (NL) QA dataset. 
Specifically, we design a set of templated NL question patterns~(template details can be found in Appendix~\ref{app:qa_template}) and instantiate the network requirements to semantically grounded, positive factual questions in NL. 
To ensure balanced evaluation, we additionally generate negative instances by introducing invalid requirements with respect to a given configurations, which are subsequently validated through manual verification to confirm their incorrectness under the given configuration. 
This process yields a comprehensive QA benchmark containing both positive and negative questions in the ratio of $1:1$.
For each requirement scale ($\{1,2,4,8,16\}$), requirement type (ExistPath, OrderedPath, LoadBalance, KConnected), and network scale (Small, Medium, Large), we generate 5 positive (and 5 negative) samples. 
In total, the benchmark contains $5 \times 4 \times 3 \times 5 \times 2 = 600$ QA pairs.



\subsection{Experiment Setting}
\paragraph{Model and Hyperparameter Settings.}
We evaluate our approach and baselines using SOTA open-weight language models, specifically the Qwen~\cite{yang2025qwen3} and Llama families~\cite{touvron2023llama}. 
Considering inference latency requirements in practical operational scenarios, we select models under 32B parameters, including Qwen2.5-32B, Qwen3-8B, Qwen3-14B, Qwen3-32B, and Llama3.1-8B, which are deployed on a server with 4 $\times$ NVIDIA A6000 GPUs.
For fact extraction, we fine-tune Llama3.1-8B using supervised fine-tuning~(SFT)~\cite{dong2024abilities} to enhance its ability to map heterogeneous and vendor-specific configurations into an unified fact graph IR. 
The SFT dataset consists of 471 manually annotated configuration-to-fact pairs curated by domain experts to ensure correctness.
We employ low-rank adaptation (LoRA)~\cite{hu2022lora} for parameter-efficient fine-tuning. 
The rank is set to 16, with an initial learning rate of $2\times 10^{-4}$, and a training duration of 100 steps using a batch size of 8. 

\input{tabs/compare}

\input{tabs/fact_acc}
\paragraph{Evaluation Metrics.}
For end-to-end performance evaluation, we use \textbf{Precision}, \textbf{Recall}, \textbf{F1} and code generation \textbf{Latency} metrics computed on the QA benchmark.
To evaluate the performances of facts extraction, we adopt a \textbf{Graph Match Ratio}~(GMR) metric that quantifies the structural and semantic alignment between the predicted fact graph $G_1$ with the ground truth graph $G_2$ annotated by human.
Formally, it can be defined as follow:
\begin{align}
\text{GMR} &= \frac{C_V + C_E}{|V_2| + |E_2|},\\
C_V &= \sum_{u \in V_1} \mathbf{1}\big(D(u) \neq \bot \land \alpha(u) = \alpha'(D(u))\big), \nonumber\\
C_E &= \sum_{(u,v) \in E_1} \mathbf{1}\big((D(u), D(v)) \in E_2 \nonumber\\&\land \beta((u,v)) = \beta'((D(u),D(v)))\big), \nonumber
\end{align}
where $ V_1 $ and $ V_2 $ denote the node sets, $ E_1 $ and $ E_2 $ denote the edge sets, $ \alpha $ and $ \alpha' $ are the node attribute functions, and $ \beta $ and $ \beta' $ are the edge attribute functions of $ G_1 $ and $ G_2 $, respectively; $ D: V_1 \to V_2 \cup \{\bot\} $ is the node mapping function where $ \bot $ indicates no match; $ \mathbf{1}(\cdot) $ is the indicator function that returns 1 if the corresponding structural and attribute match is correct and 0 otherwise; and the denominator $ |V_2| + |E_2| $ is the total number of nodes and edges in $ G_2 $.

\subsection{End-to-End Performance of \ours}
In this part, we evaluate \ours's end-to-end performances on answering NL network questions.
Specifically, we compare against two baseline approaches with various LLM backbones, and report Precision, Recall, and F1 scores.
The first baseline, \textit{Direct QA}, directly feeds all router configurations into the LLM to produce answers.
The second baseline follows a Retrieval-Augmented Generation (RAG) paradigm~\cite{lewis2020retrieval}: given the large size of configuration files, we apply the commonly used fixed-length chunking~\cite{wang2025document} to split them into snippets, embed them using sentence transformer~\cite{reimers1908sentence}, and retrieve the most relevant snippets to support answer generation.

As the comparison results shown in Table~\ref{tab:compare}, Direct QA performs the worst, with performance degrading as network size grows.
This decline stems from the difficulty of reasoning over complex protocols within lengthy configurations, and from context-length limitations when handling large-scale networks.
The RAG-based method can alleviate the long-context limitation and achieves improved performance, but still falls short of \ours.
Among \ours's variants, Qwen3-8B outperforms Llama3.1-8B in most cases, reflecting its stronger reasoning capabilities acquired through reinforcement learning in post-training~\cite{yang2025qwen3}. 
Within the Qwen family, performance improves with larger model sizes, highlighting the scalability potential of \ours on larger LLMs.
Qwen3-32B yields the highest average F1 (around 85\%), demonstrating that \ours is a promising framework for real-world network management with potential to reduce operational expense. 
Notably, unlike Direct QA and RAG-based QA, \ours does not suffer performance degradation as network scale increases, since it generates executable solution code rather than attempting direct inference, thereby guarantees the scalability of \ours on large networks. 
Note that for fair comparison, all \ours variants employ the same fact extraction model, \ie Llama3.1-8B-SFT as described in \S\ref{sec:fgir}.

\subsection{Effectiveness of Tree-Based Chunking}
The tree-based chunking method introduced in \S\ref{sec:chunk} is designed to preserve semantic integrity while mitigating the understanding and reasoning burden imposed by long-context inputs on LLMs.
Table~\ref{tab:compare} compares the performances of \ours using tree-based chunking method with Direct QA~(without chunking) and RAG-based QA~(fixed-length chunking) methods.
As previously discussed, long-context configurations bring understanding and reasoning challenges for LLMs, leading to performance degradation.
Consistent with this observation, Direct QA achieves the lowest performance. 
Moreover, as network scale increases, most configurations surpass the maximum input length supported by LLMs, resulting in complete failure (\ie zero precision and recall). 
While RAG-based QA can alleviate this issue, the fixed-length chunking strategy often disrupts semantic coherence ~(see the case study of chunking methods in Appendix~\S\ref{app:chunk}), thereby preventing the model from accurately reasoning about network behaviors.
In contrast, \ours leverages tree-based chunking strategy to decompose lengthy configurations into semantically coherent paths.
This approach avoids information fragmentation and preserves the contextual relationships necessary for accurate reasoning. 
Results in Table~\ref{tab:compare} confirm that \ours significantly outperforms both baselines, demonstrating the efficacy of structure-aware chunking for long-context network configuration understanding and reasoning.

\begin{figure}[tbp]
    \centering
    \captionsetup[subfigure]{labelformat=parens}
    
    \subfloat[ExistPath]{
        \includegraphics[width=0.45\columnwidth]{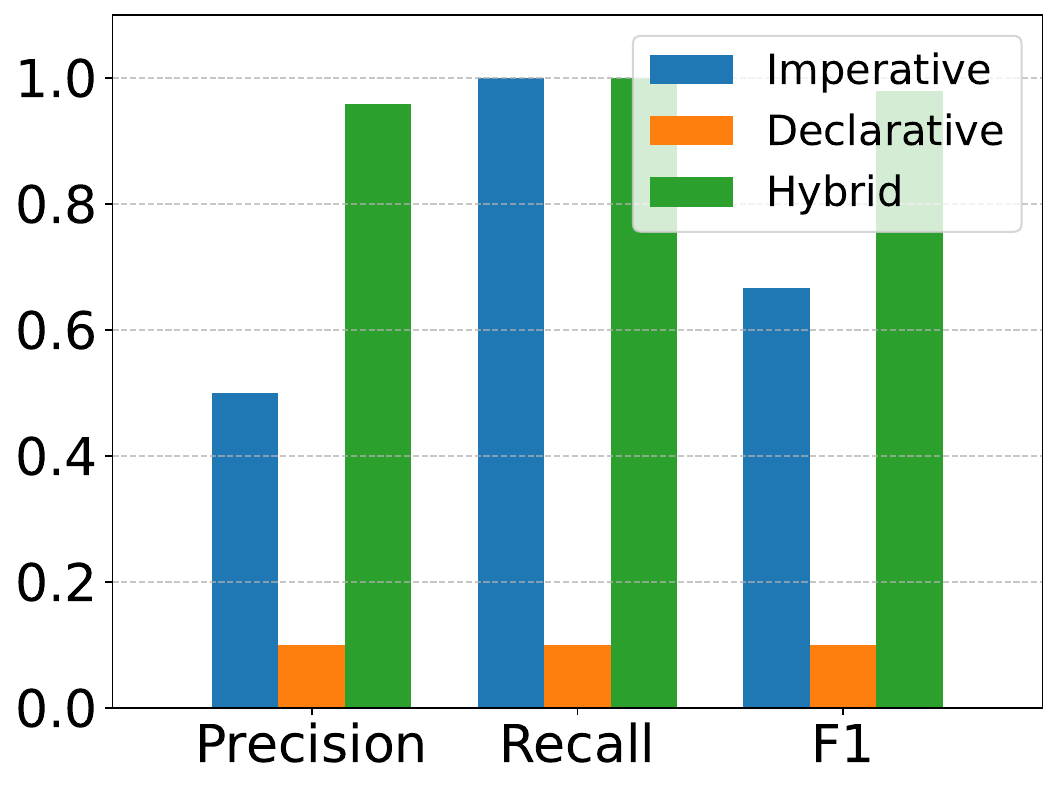}
        \label{fig:simple}
    }
    \hfill
    \subfloat[OrderedPath]{
        \includegraphics[width=0.45\columnwidth]{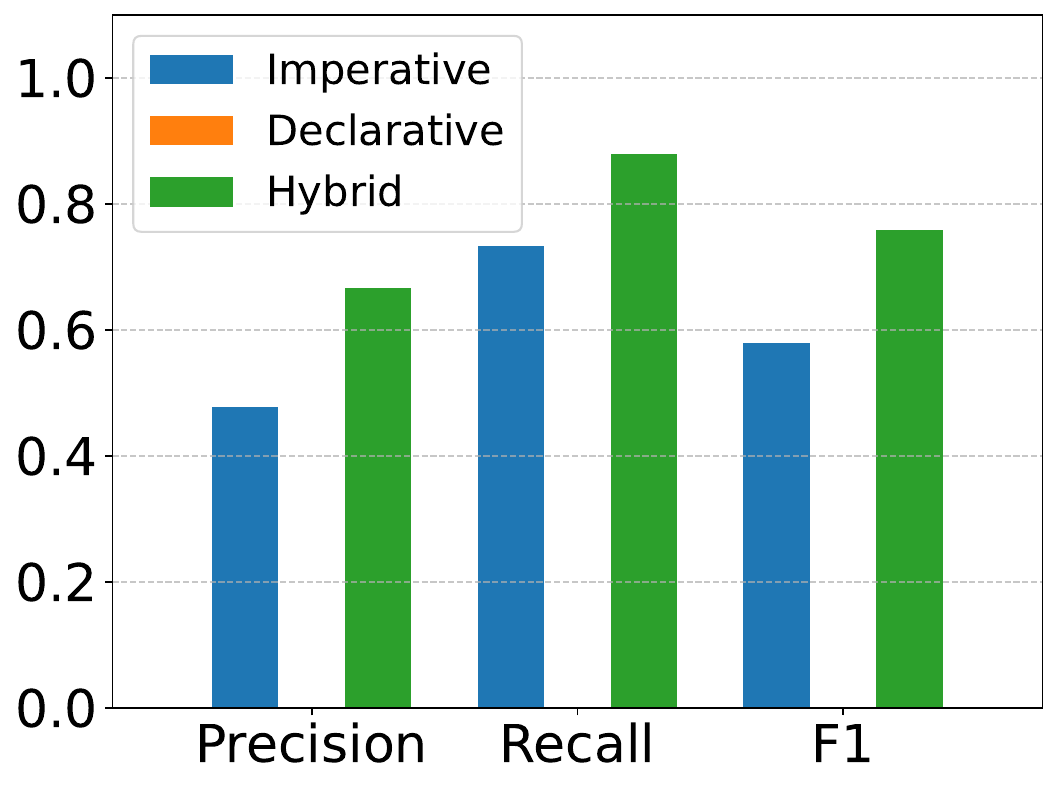}
        \label{fig:order}
    }
    
    
    \subfloat[LoadBalance]{
        \includegraphics[width=0.45\columnwidth]{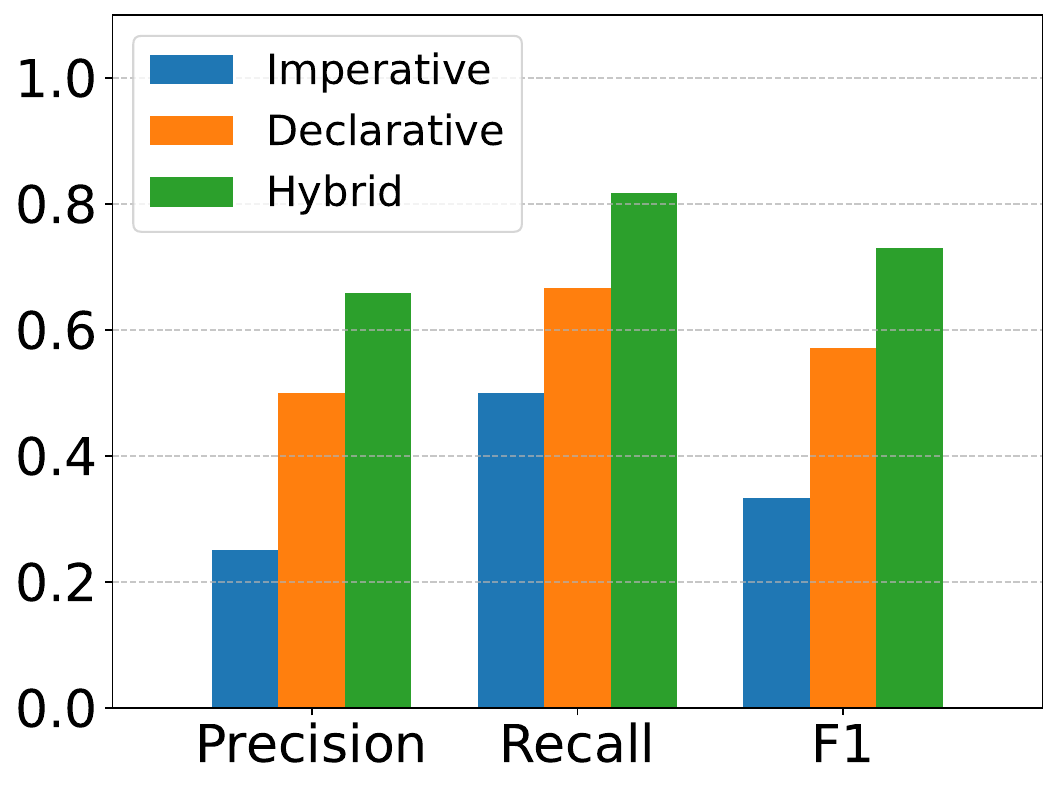}
        \label{fig:ecmp}
    }
    \hfill
    \subfloat[KConnected]{
        \includegraphics[width=0.45\columnwidth]{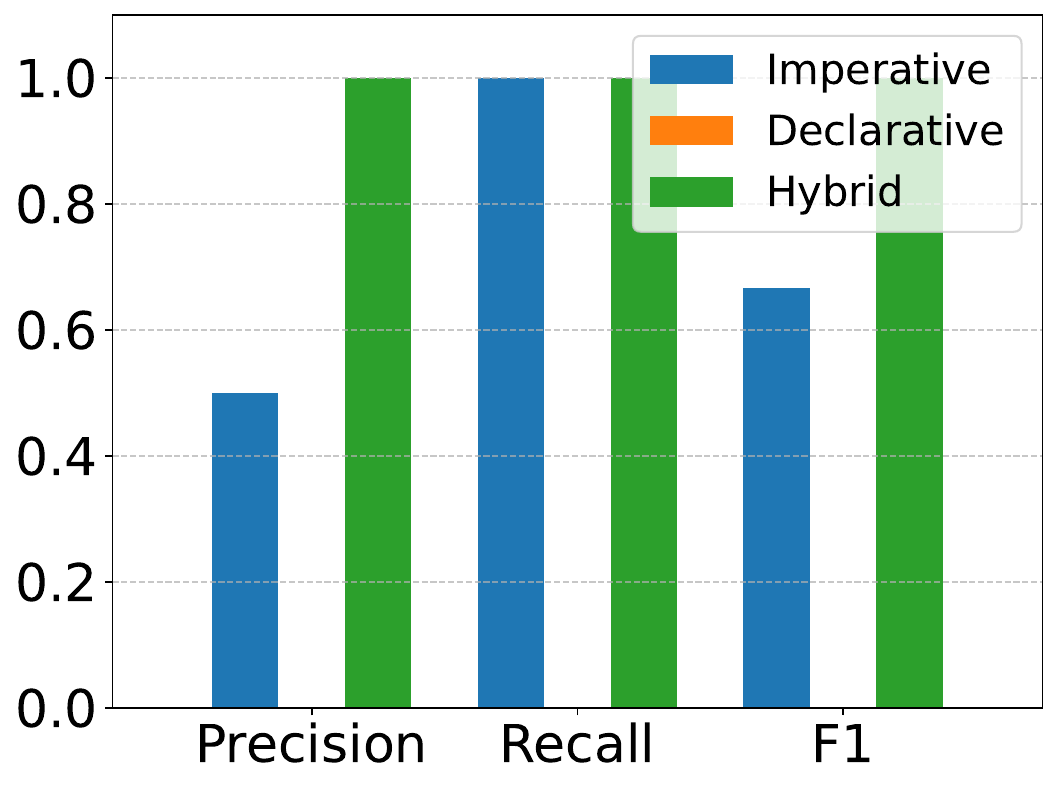}
        \label{fig:kconnected}
    }
    
    \caption{Comparisons of imperative, declarative and hybrid languages on answering four types of questions.}
    \label{fig:code}
\end{figure}

\subsection{GMR of Fact Graph Transformation}
We design explicit fact extraction and implicit fact deduction methods in \S\ref{sec:fgir} to accurately and comprehensively capture the factual knowledge of network states.
To evaluate the effectiveness of this approach, we compute the Graph Matching Ratio (GMR) between the constructed fact graph and the human-annotated ground truth, reporting results in Table~\ref{tab:fact_acc}.
Interestingly, larger Qwen models do not necessarily yield better extraction accuracy in this task.
Upon reading their output, we observe that overly elaborate reasoning often introduces errors, whereas a moderate reasoning depth achieves higher GMR.
Among the baseline LLMs, Llama3.1-8B attains the best average performance. 
Moreover, fine-tuning Llama3.1-8B on annotated configuration–to-fact pairs yields \textit{Llama3.1-8B-SFT}, which significantly outperforms all baseline models.

\begin{figure}[tbp]
    \centering
    \captionsetup[subfigure]{labelformat=parens}
    
    \subfloat[Precision]{
        \includegraphics[width=0.45\columnwidth]{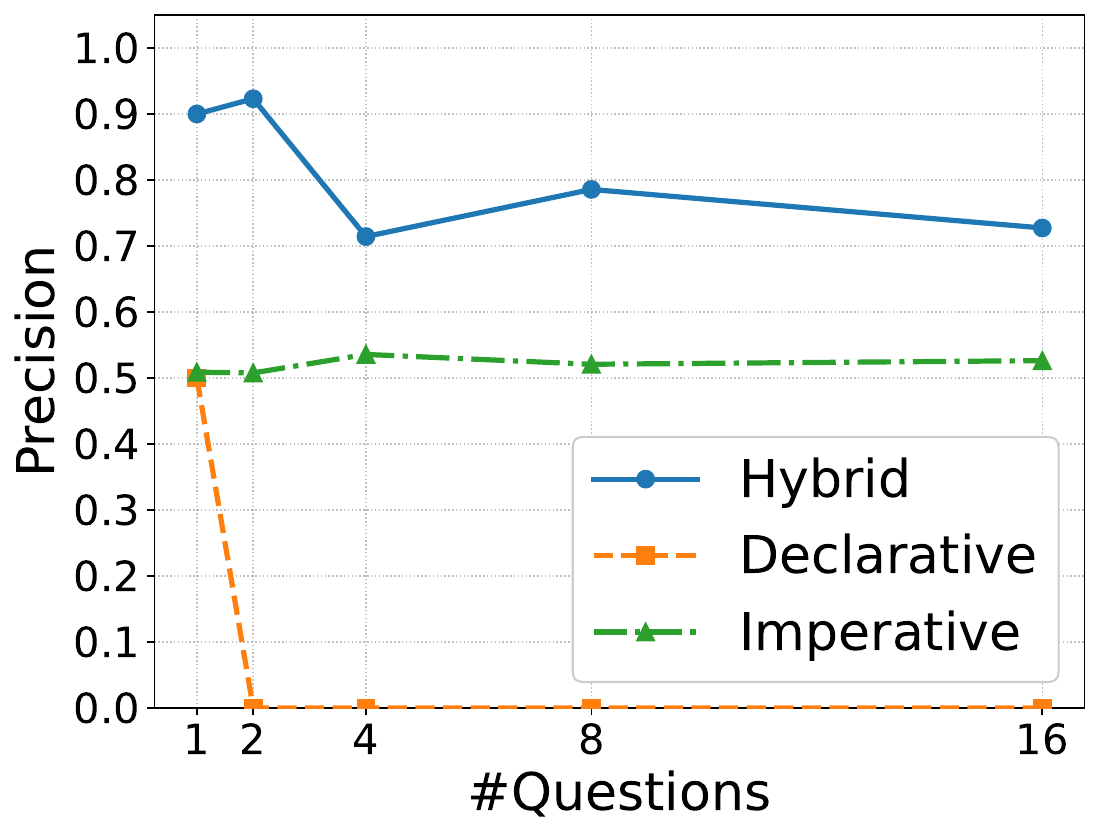}
        \label{fig:q_preicison}
    }
    \hfill
    \subfloat[Recall]{
        \includegraphics[width=0.45\columnwidth]{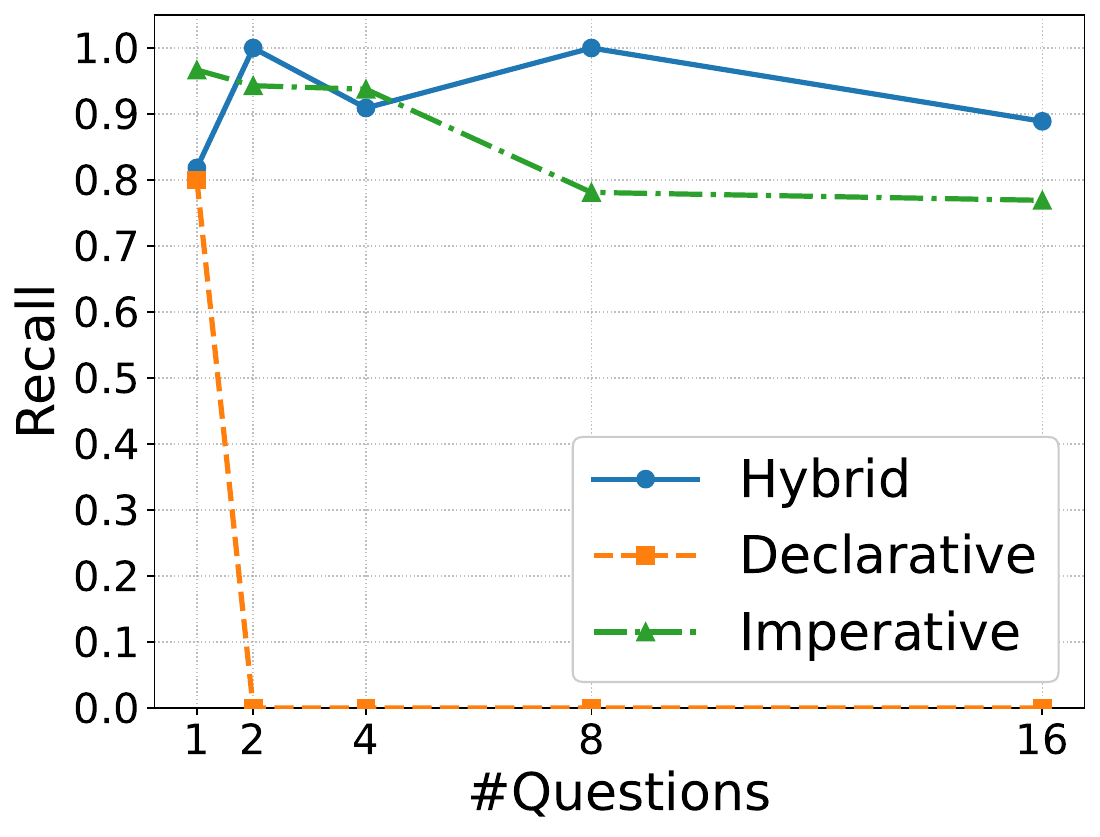}
        \label{fig:q_recall}
    }
    
    
    \subfloat[F1]{
        \includegraphics[width=0.45\columnwidth]{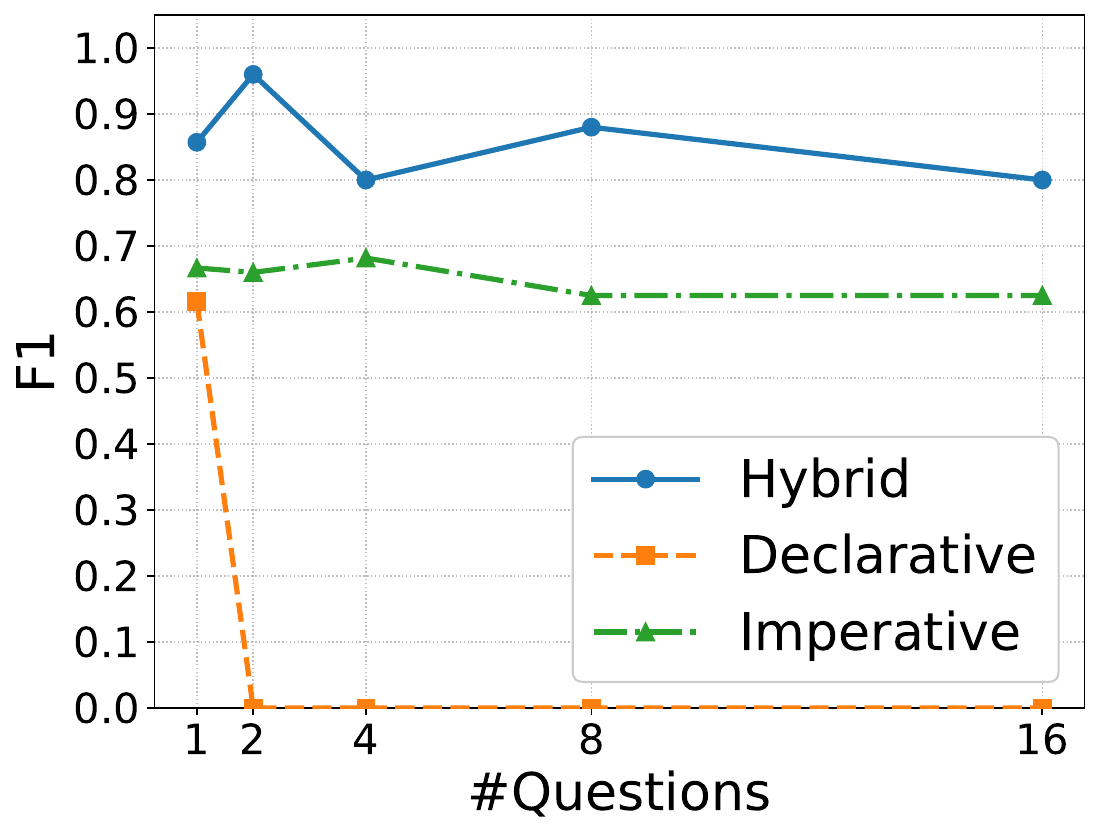}
        \label{fig:q_f1}
    }
    \hfill
    \subfloat[Latency]{
        \includegraphics[width=0.45\columnwidth]{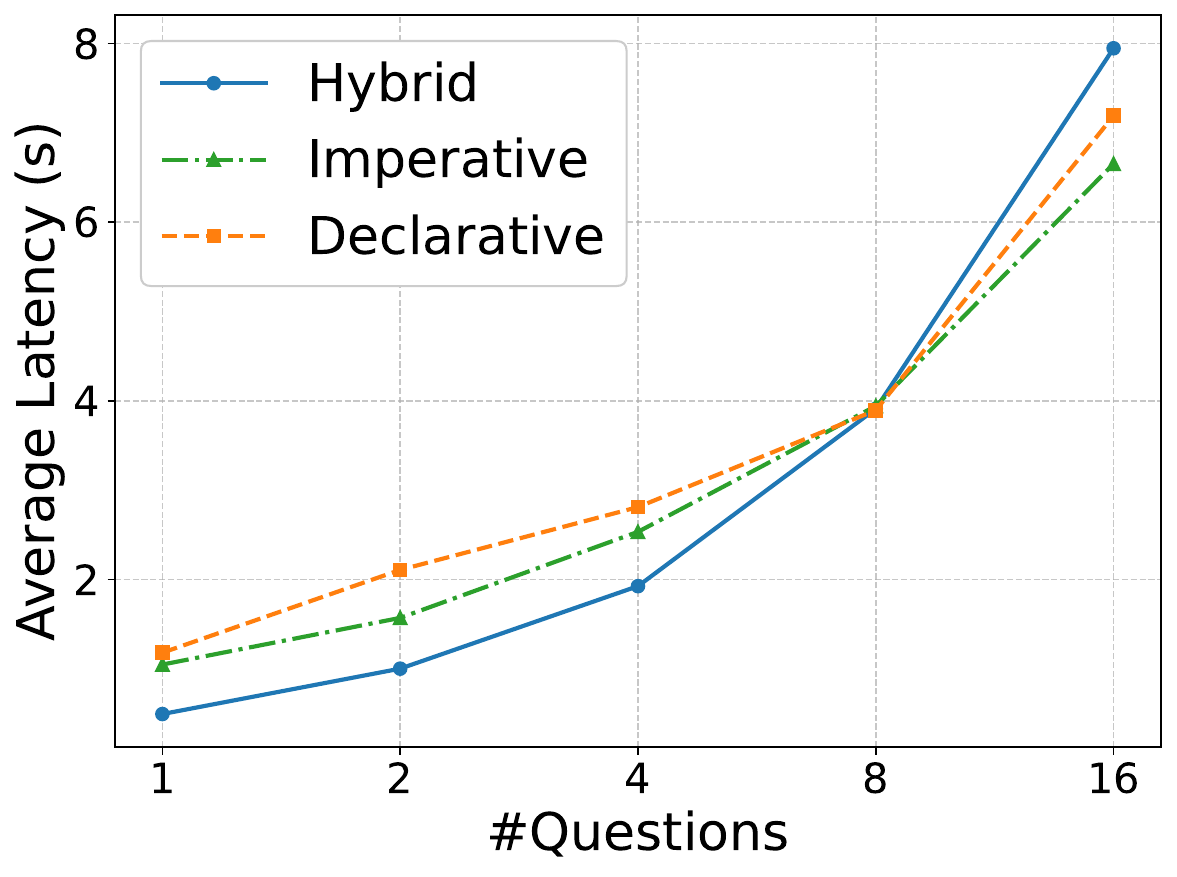}
        \label{fig:q_latency}
    }
    
    \caption{Comparisons of imperative, declarative and hybrid languages on answering different number of questions.}
    \label{fig:q_num}
\end{figure}

\subsection{Comparisons of Hybrid Language}
In \S\ref{sec:hybrid_code}, we propose a hybrid language integrating the simplicity of declarative representations and the expressive reasoning capability of imperative programming.
To assess it, we compare against two baselines: 
(i) an imperative baseline that uses Python (with access to standard packages such as NetworkX~\cite{hagberg2008exploring}) to directly answer questions, 
and (ii) a declarative baseline that generates Batfish~\cite{fogel2015general} programs, a Datalog-based network verification tool.

Fig.~\ref{fig:code} reports Precision, Recall, and F1 scores across four types of network questions using Qwen3-32B as the backbone (results for other LLMs are similar and omitted for brevity). 
We observe that the declarative approach succeeds only on \textit{ExistPath} and \textit{LoadBalance} set, which require limited reasoning. 
For more complex questions such as \textit{OrderedPath} and \textit{KConnected}, which demand logical and numerical computation, the declarative method fails to produce correct results. 
In contrast, recent LLMs such as Qwen demonstrate strong proficiency in generating imperative Python programs~\cite{jiang2024survey}, which, combined with NetworkX graph algorithms, enable the imperative baseline to handle reasoning tasks. 
Our method, \ours, consistently outperforms both baselines, validating the benefit of combining declarative succinctness with imperative reasoning.
 
We further vary the number of questions and report the corresponding performance metrics (Precision, Recall, F1) and latency in Fig.~\ref{fig:q_num}. 
Results indicate that the declarative baseline fails once the question number exceeds 2, while the imperative approach maintains stable performance across all scales. 
Notably, \ours outperforms both baselines across nearly all settings, highlighting the scalability and effectiveness of the proposed hybrid language.
Moreover, the hybrid language exhibits lower latency due to its concise representation, except when the number of questions reaches 16, where more logical constraints are introduced to ensure answer correctness.
A detailed case analysis of different generated codes is provided in Appendix~\S\ref{app:code}.

%% file: tabs/compare.tex
\begin{table*}[tbp]
\centering
\caption{The end-to-end QA performance across different methods, LLMs and network scales. The best and second results are marked in {bold} and {underline}, respectively.}
\label{tab:compare}
\begin{tabular}{ccccccccccc}
\toprule
\multicolumn{2}{c}{Network Scale}           & \multicolumn{3}{c}{Small}   & \multicolumn{3}{c}{Medium}  & \multicolumn{3}{c}{Large}   \\
\midrule
\multicolumn{2}{c}{Metrics}                 & Precision & Recall & F1     & Precision & Recall & F1     & Precision & Recall & F1     \\
\midrule
\multirow{5}{*}{Direct QA}    & Llama3.1-8B & 0.4966    & 0.8231 & 0.6195 & 0.5000    & \underline{0.8889} & 0.6400 & 0.4954    & \underline{0.9310} & 0.6467 \\
                              & Qwen3-8B    & 0.5057    & 0.5714 & 0.5366 & 0.7333    & 0.4167 & 0.5314 & 0.0000    & 0.0000 & 0.0000 \\
                              & Qwen3-14B   & 0.5667    & 0.2208 & 0.3178 & 0.0000    & 0.0000 & 0.0000 & 0.0000    & 0.0000 & 0.0000 \\
                              & Qwen2.5-32B & 0.6571    & 0.6133 & 0.6345 & 0.5000    & 0.1000 & 0.1667 & 0.0000    & 0.0000 & 0.0000 \\
                              & Qwen3-32B   & 0.6429    & 0.3506 & 0.4538 & 0.0000    & 0.0000 & 0.0000 & 0.0000    & 0.0000 & 0.0000 \\
\midrule                              
\multirow{5}{*}{RAG-based QA} & Llama3.1-8B & 0.5152    & 0.8718 & 0.6476 & 0.5357    & 0.8333 & 0.6522 & 0.5053    & {0.8276} & 0.6275 \\
                              & Qwen3-8B    & 0.5957    & 0.3590 & 0.4480 & 0.5714    & 0.2222 & 0.3200 & 0.5517    & 0.2759 & 0.3678 \\
                              & Qwen3-14B   & 0.6364    & 0.0897 & 0.1573 & \textbf{1.0000}    & 0.0556 & 0.1053 & 0.0000    & 0.0000 & 0.0000 \\
                              & Qwen2.5-32B & 0.6154    & 0.4103 & 0.4923 & 0.6250    & 0.2778 & 0.3846 & \underline{0.7097}    & 0.3793 & 0.4944 \\
                              & Qwen3-32B   & 0.5294    & 0.1154 & 0.1895 & \textbf{1.0000}    & 0.0556 & 0.1053 & 0.5714    & 0.0690 & 0.1231 \\
\midrule
\multirow{5}{*}{NetMind QA}   & Llama3.1-8B & 0.3333    & 0.6000 & 0.4286 & 0.3333    & \textbf{1.0000} & 0.5000 & 0.6667    & \textbf{1.0000} & 0.8000 \\
                              & Qwen3-8B    & 0.4348    & 0.6667 & 0.5263 & 0.6000    & \textbf{1.0000} & 0.7500 & 0.6667    & 0.5000 & 0.5714 \\
                              & Qwen3-14B   & \underline{0.6585}    & 0.7297 & 0.6923 & 0.7500    & \textbf{1.0000} & \underline{0.8571} & 0.7000    & \textbf{1.0000} & \underline{0.8235} \\
                              & Qwen2.5-32B & 0.5690    & \textbf{0.9706} & \underline{0.7174} & \underline{0.7778}    & \textbf{1.0000} & \textbf{0.8750} & 0.5000    & 0.7500 & 0.6000 \\
                              & Qwen3-32B   & \textbf{0.8500}    & \underline{0.8947} & \textbf{0.8718} & 0.7000    & \textbf{1.0000} & 0.8235 & \textbf{0.7273}    & \textbf{1.0000} & \textbf{0.8421} \\
\bottomrule
\end{tabular}
\end{table*}

%% file: tabs/fact_acc.tex
\begin{table}[tbp]
\centering
\caption{The Graph Match Ratio of fact graph IR.}
\label{tab:fact_acc}
\begin{tabular}{ccccc}
\toprule
                & Small  & Medium & Large  & Avg.     \\
\midrule
Llama3.1-8B     & 0.6763 & 0.4854 & 0.1864 & 0.4494   \\
Qwen3-8B        & 0.4595 & 0.1933 & 0.2265 & 0.2931   \\
Qwen3-14B       & 0.5278 & 0.0000 & 0.0000 & 0.1759   \\
Qwen2.5-32B     & 0.0334 & 0.0000 & 0.0000 & 0.0111   \\
Qwen3-32B       & 0.4916 & 0.0000 & 0.0000 & 0.1639 \\ \midrule
Llama3.1-8B-SFT & \textbf{0.8925} & \textbf{0.8042} & \textbf{0.8489} & \textbf{0.8485}  \\
\bottomrule
\end{tabular}
\end{table}

%% file: sections/06-conclusion.tex
\section{Conclusion}
In this paper, we present \ours, a novel framework for understanding network behaviors through answering natural language questions. 
While LLMs exhibit strong understanding and reasoning capabilities in network scenarios, we identify three key challenges in this task and introduce tailored modules to address them.
Specifically, we propose a tree-based chunking method to reduce input context while preserving semantic coherence, a unified fact graph IR to model heterogeneous configurations, and a hybrid imperative–declarative language to enable accurate reasoning.
Extensive experiments conducted on a newly constructed benchmark validate the effectiveness of \ours, showing consistent improvements over strong baselines. 
In future work, we plan to incorporate reinforcement learning to further enhance the reasoning and code generation capabilities of \ours, thereby extending its applicability to real-world network management scenarios.

%% file: sections/07-appendix.tex
\section{Appendix}



\subsection{Fact Base and Rule Set}
\label{app:facts}
From raw network configurations, we extract a set of facts representing factual knowledge about the network states.
Table~\ref{tab:fact_base} provides the complete fact base and corresponding descriptions.

\begin{table}[h]
    \centering
        \caption{Fact base and corresponding descriptions.}
    \begin{tabular}{ll}
    \toprule
        Fact & Description \\ \midrule
        Router($r$) & a router with hostname $r$\\
        Interface($i$, $IP$/$Mask$) & \makecell[l]{an interface named $i$ with ip \\address $IP$ and subnet mask $Mask$}\\
        IPPrefixList($n$, $s$, $a$, $IP$/$Mask$) & \makecell[l]{an IP prefix list named $n$ with\\ sequence number $s$, access $a$, ip\\ address $IP$ and subnet mask $Mask$}\\
        CommunityList($i$, $a$, [$c$]) & \makecell[l]{a list of communities with ID $i$,\\ access $a$,\\ and a list of community names $c$}\\
        RouteMap($n$, $s$, $a$, $m$, $n$) & \makecell[l]{a route policy named $n$, \\with sequence number $s$, access $a$. \\When $m$ is matched, \\the action $a$ is executed}\\
        BGP($asn1$, $n$($IP$, $asn2$, $r$, $in$, $out$)) & \makecell[l]{BGP configured for AS $asn1$, \\BGP neighbor with ip address $IP$, \\AS number $asn2$, \\router name $r$, \\inbound/outbound policy\\ $in$ and $out$}\\
        OSPFNetwork($r$, [$n$]) &  \makecell[l]{OSPF network for router $n$'s\\ interfaces [$n$]} \\
        OSPFCost($r1$, $r2$, $c$) &  \makecell[l]{OSPF cost between router\\ $r1$ and $r2$} \\
        \bottomrule
    \end{tabular}
    \label{tab:fact_base}
\end{table}

While most facts can be explicitly extracted, certain facts are implicit.
By applying predefined rules to the extracted facts, new facts can be deduced, enabling a more comprehensive understanding of network behaviors.
The complete rule set is defined as follows:
\begin{align}
    \texttt{RouteEdge}(R_1, R_2) \leftarrow
 &\land\ \texttt{Interface}(R_1, \text{IP}_1/\text{Mask}), \nonumber\\
    & \land\ \texttt{Interface}(R_2, \text{IP}_2/\text{Mask}), \nonumber\\
    & \land \ 	\lnot\ R_1 == R_2, \nonumber\\
    & \land\ \texttt{Subnet}(\text{IP}_1/\text{Mask}) == \texttt{Subnet}(\text{IP}_2/\text{Mask}). \nonumber
\end{align}

\begin{align}
    \texttt{BGPPeer}(R_1, R_2) \leftarrow
    &\land\ \texttt{BGP}(ASN_1, \texttt{Neighbors}_1),\nonumber\\
    &\land\ \texttt{member}(\texttt{Neighbor}_1, \texttt{Neighbors}_1),\nonumber\\
    &\land\ \texttt{Neighbor}_1 = \texttt{neighbor}(IP_1, ASN_2, R_2),\nonumber\\
    &\land\ \texttt{BGP}(ASN_2, \texttt{Neighbors}_2),\nonumber\\
    &\land\ \texttt{member}(\texttt{Neighbor}_2, \texttt{Neighbors}_2),\nonumber\\
    &\land\ \texttt{Neighbor}_2 = \texttt{neighbor}(IP_2, ASN_1, R_1),\nonumber\\
    &\land\ \left(\texttt{Interface}(R_2, IP_1) \lor \texttt{Interface}(R_2, IP_2)\right),\nonumber\\
    &\land\ \left(\texttt{Interface}(R_1, IP_2) \lor \texttt{Interface}(R_1, IP_1)\right),\nonumber\\
    &\land\ \lnot\ R_1 == R_2.\nonumber
\end{align}

\subsection{Datalog Predicates of Hybrid Language}
\label{app:bnf}
In the proposed hybrid language, Datalog is responsible for declarative reasoning.
Each Datalog predicate encodes the logical formulation of a typical network query in a declarative, query-like manner.
For example, $\texttt{ShortestOSPFPath}$ implements the shortest path algorithm and return answer for the given start and target routers.
The supported Datalog predicates is summarized in Table~\ref{tab:datalog_predicate}.

\begin{table}[h]
    \centering
    \caption{Supported Datalog predicates.}
    \begin{tabular}{ll}
    \toprule
      Predicate   &  Description\\ \midrule
      Reach($r1$, $r2$)   &  check if $r1$ can directly reach $r2$\\
      ExistPath([$r$])  & check if exist a route path along [$r$]\\
      AllPath($r1$, $r2$)  & return all route paths from $r1$ to $r2$\\
      TraceRoute($r1$, $r2$)  & \makecell[l]{return the default route path\\ from $r1$ to $r2$}\\
      ShortestOSPFPath($r1$, $r2$) & \makecell[l]{return the path with the \\lowest OSPF cost from $r1$ to $r2$}\\
      OSPFWeight([$r$]) & \makecell[l]{return the summarization of \\OSPF weights along the path [$r$]}\\
      \bottomrule
    \end{tabular}
    
    \label{tab:datalog_predicate}
\end{table}

\subsection{Question Templates for Natural Language}
\label{app:qa_template}
To generate questions represented in natural language, we design a set of QA templates, which are specified as follows:
\begin{tcolorbox}[breakable=true, colback=blue!5!white, colframe=blue!70!black, title=ExistPath Templates]
Is \{path\} a valid route?\\
Can traffic flow through \{path\}?\\
Does this route path exist: \{path\}?\\
Does the network permit the route path \{path\}?
\end{tcolorbox}

\begin{tcolorbox}[breakable, colback=blue!5!white, colframe=blue!70!black, title=OrderedPath Templates]
For the traffic from \{src\} to \{dst\}, does the route path \{path1\} be prioritized instead of \{path2\}?\\
When routing traffic from \{src\} to \{dst\}, does the route path \{path1\} be favored instead of \{path2\}?\\
For optimal routing from \{src\} to \{dst\}, does the route path \{path1\} be selected over \{path2\}?\\
For the traffic from \{src\} to \{dst\}, does the route path \{path1\} should always be chosen over \{path2\}?\\
Does the primary route path for the traffic from \{src\} to \{dst\} should be \{path1\}? The \{path2\} is a fallback only.\\
Does the traffic from \{src\} to \{dst\} should prefer the route path \{path1\} over \{path2\}?
\end{tcolorbox}

\begin{tcolorbox}[breakable, colback=blue!5!white, colframe=blue!70!black, title=LoadBalance Templates]
Check if \{path1\} and \{path2\} are load-balanced due to equal-cost routing.\\
Are the routing preferences of paths \{path1\} and \{path2\} the same?\\
Do paths \{path1\} and \{path2\} have equal routing preferences?\\
Are \{path1\} and \{path2\} configured with identical routing priorities?\\
Does \{path1\} share the same routing preference as \{path2\}?\\
Does the routing protocol assign equal preference to \{path1\} and \{path2\}?
\end{tcolorbox}

\begin{tcolorbox}[breakable, colback=blue!5!white, colframe=blue!70!black, title=KConnected Templates]
Are there two paths connecting \{start\} with \{end\}, i.e., paths \{path1\} and \{path2\}?\\
Does the network provide two separate routes from \{start\} to \{end\}, i.e., paths \{path1\} and \{path2\}?\\
Can traffic flow from \{start\} to \{end\} via two different paths, i.e., paths \{path1\} and \{path2\}?\\
Whether \{start\} and \{end\} are connected by two paths, i.e., paths \{path1\} and \{path2\}?\\
For fault tolerance, are there two backup paths from \{start\} to \{end\}, i.e., paths \{path1\} and \{path2\}?
\end{tcolorbox}

\subsection{Prompts for Fact Extraction and Code Generation}
\label{app:prompt_template}

We use LLMs to extract facts from configurations and generate code.
We present the prompts used as follows~(the complete prompts can be found in our released code):

\begin{tcolorbox}[breakable, colback=blue!5!white, colframe=blue!70!black, title=Fact Extraction Prompt]
You are an expert in parsing router configuration files into the following predefined facts, where the content after `@` denotes the data type of the corresponding parameter, and the content after `\#` is the comments on the facts:\\
add\_router(ROUTER\_NAME@String, interfaces=[(INTERFACE\_1@String, IP\_1@String, PEER\_ROUTER\_NAME\_1@String), (INTERFACE\_2@String, IP\_2@String, PEER\_ROUTER\_NAME\_2@String), ...])\\
......\\
......\\
Each line or group of lines in the configuration file must be mapped to one of the above facts. If no corresponding facts is found, the line should be ignored.
\end{tcolorbox}

\begin{tcolorbox}[breakable, colback=blue!5!white, colframe=blue!70!black, title=Hybrid Code Generation Prompt]
You are an AI expert specializing in network analysis. Your task is to generate mixed Python and Datalog programs to answer network-related questions. Python is used for imperative logic and complex reasoning, while Datalog is utilized for declarative querying of network properties.\\
The available datalog predicates are as follows:\\
- ExistPath([Router1, Router2, ..., RouterN]): Returns True if the specified sequence of routers [Router1, Router2, ..., RouterN] forms a valid path.\\
......\\
......\\
You will be provided with one or more network questions. Generate a mixed program to answer them.\\
Use native Python syntax for the imperative parts.\\
Only use the above Datalog predicates for the declarative parts.
\end{tcolorbox}

\subsection{Case Analysis of Chunking Algorithms}
\label{app:chunk}
\begin{figure}[tbp]
    \centering
    \includegraphics[width=\columnwidth]{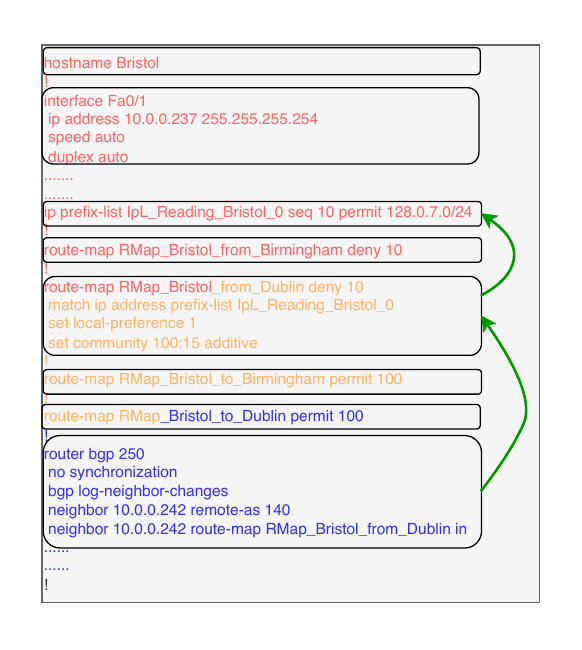}
    \caption{Comparisons of different chunking algorithms. The fixed-length chunking splits configuration into broken parts~(in red, yellow and blue). In contrast, our proposed tree-based chunking can preserve the semantic dependencies and coherence~(in black boxes and green curves).}
    \label{fig:app_chunk_example}
\end{figure}

Conventional fixed-length chunking method often disrupt the semantic integrity of network configurations, leading to incomplete or fragmented interpretations.
In contrast, our proposed tree-based chunking approach preserves semantic coherence by leveraging structural dependencies within the configuration.
Due to page limit, we illustrate a small case to compare the two methods in Fig.~\ref{fig:app_chunk_example}.
As shown, the fixed-length chunking incorrectly divides the $\texttt{RouteMap}$ block, resulting in inaccurate reasoning.
Conversely, tree-based chunking first identifies semantically meaningful configuration blocks and then links them through a dependency tree.
By traversing the tree along root-to-leaf paths, our approach effectively preserves both syntactic structure and semantic coherence.

\subsection{Case Analysis of Hybrid Language}
\label{app:code}

We integrate the advantages of both imperative and declarative paradigms to design a hybrid language tailored for network reasoning tasks.
This language enables accurate representation of complex reasoning logic while alleviating the reasoning burden on LLMs.
To demonstrate its effectiveness, we present a case study involving the following questions:

\textit{(1) For traffic from Ajdovščina to Divača, is the direct route ['Ajdovščina', 'Divača'] prioritized over ['Ajdovščina', 'NovaTTGorica', 'Ljubljana', 'Koper', 'Divača']?\\
\indent (2) For traffic from Divača to Ajdovščina, is the route ['Divača', 'Ajdovščina'] preferred over ['Divača', 'Senožeče', 'Postojna', 'Ljubljana', 'NovaTTGorica', 'Ajdovščina']?}

By comparing the three code variants, we observe that while the purely imperative program is syntactically correct, it focuses solely on computing the shortest physical path rather than the actual routing path, leading to incorrect results.
The purely declarative program, on the other hand, introduces hallucinated Datalog-like queries that fail during execution.
In contrast, our hybrid approach abstracts and delegates routine, query-oriented computations to the declarative component while employing imperative logic for higher-level reasoning.
This design paradigm proves more compatible with LLMs, effectively reducing their reasoning load and improving execution reliability.

\begin{tcolorbox}[breakable, colback=blue!5!white, colframe=blue!70!black, title=Imperative Code]
\begin{lstlisting}[language=Python]
[
breaklines=true,
breakanywhere=true,
]{python}
import networkx as nx
path1_1 = ['Ajdovscina', 'Divaca']
path1_2 = ['Ajdovscina', 'NovaTTGorica', 'Ljubljana', 'Koper', 'Divaca']
path2_1 = ['Divaca', 'Ajdovscina']
path2_2 = ['Divaca', 'Senozece', 'Postojna', 'Ljubljana', 'NovaTTGorica', 'Ajdovscina']
direct_path_exists = nx.has_path(g, path1_1[0], path1_1[-1])
direct_path_length = len(path1_1)
longer_path_exists = nx.has_path(g, path1_2[0], path1_2[-1])
longer_path_length = len(path1_2)
is_direct_path_shorter = direct_path_length < longer_path_length
shortest_path = nx.shortest_path(g, path1_1[0], path1_1[-1])
is_direct_path_shortest = shortest_path == path1_1
shortest_path2 = nx.shortest_path(g, path2_1[0], path2_1[-1])
is_direct_path_shortest2 = shortest_path2 == path2_1
answer = is_direct_path_shortest and is_direct_path_shortest2
% \end{minted}
\end{lstlisting}
\end{tcolorbox}

\begin{tcolorbox}[breakable, colback=blue!5!white, colframe=blue!70!black, title=Declarative Code]
\begin{lstlisting}[language=Python]
[
breaklines=true,
breakanywhere=true,
]{python}
from pybatfish.client.session import Session
from pybatfish.question import bfq
bf = Session(host="localhost")
bf.set_network("example_network")
bf.init_snapshot("CONFIG_FOLDER", name=f"snapshot", overwrite=True)
source1 = "Ajdovscina"
dest1 = "Divaca"
alt_path1 = ["Ajdovscina", "NovaTTGorica", "Ljubljana", "Koper", "Divaca"]
source2 = "Divaca"
dest2 = "Ajdovscina"
alt_path2 = ["Divaca", "Senozece", "Postojna", "Ljubljana", "NovaTTGorica", "Ajdovscina"]
q1 = bfq.traceroute(src=source1, dst=dest1).answer()
preferred_path1 = q1.df().iloc[0]['hop'].split(" -> ") if not q1.df().empty else []
q2 = bfq.traceroute(src=source2, dst=dest2).answer()
preferred_path2 = q2.df().iloc[0]['hop'].split(" -> ") if not q2.df().empty else []
fact1 = preferred_path1 == [source1, dest1]
fact2 = preferred_path2 == [source2, dest2]
answer = fact1 and fact2
% \end{minted}
\end{lstlisting}
\end{tcolorbox}

\begin{tcolorbox}[breakable, colback=blue!5!white, colframe=blue!70!black, title=Hybrid Code]
\begin{lstlisting}[language=Python]
[
breaklines=true,
breakanywhere=true,
]{python}
p1 = ['Ajdovscina', 'Divaca']
p2 = ['Ajdovscina', 'NovaTTGorica', 'Ljubljana', 'Koper', 'Divaca']
p3 = ['Divaca', 'Ajdovscina']
p4 = ['Divaca', 'Senozece', 'Postojna', 'Ljubljana', 'NovaTTGorica', 'Ajdovscina']
res1 = ExistPath(p1)
res2 = ExistPath(p2)
res3 = ExistPath(p3)
res4 = ExistPath(p4)
if res1 is None or res2 is None or res3 is None or res4 is None:
    answer = False
res5 = TraceRoute('Ajdovscina', 'Divaca')
if res5 != p1:
    answer = False
res6 = TraceRoute('Divaca', 'Ajdovscina')
if res6 != p3:
    answer = False
% \end{minted}
\end{lstlisting}
\end{tcolorbox}

%% file: main.bbl

\begin{thebibliography}{57}


\ifx \showCODEN    \undefined \def \showCODEN     #1{\unskip}     \fi
\ifx \showISBNx    \undefined \def \showISBNx     #1{\unskip}     \fi
\ifx \showISBNxiii \undefined \def \showISBNxiii  #1{\unskip}     \fi
\ifx \showISSN     \undefined \def \showISSN      #1{\unskip}     \fi
\ifx \showLCCN     \undefined \def \showLCCN      #1{\unskip}     \fi
\ifx \shownote     \undefined \def \shownote      #1{#1}          \fi
\ifx \showarticletitle \undefined \def \showarticletitle #1{#1}   \fi
\ifx \showURL      \undefined \def \showURL       {\relax}        \fi
\providecommand\bibfield[2]{#2}
\providecommand\bibinfo[2]{#2}
\providecommand\natexlab[1]{#1}
\providecommand\showeprint[2][]{arXiv:#2}

\bibitem[Abhashkumar et~al\mbox{.}(2020)]%
        {abhashkumar2020tiramisu}
\bibfield{author}{\bibinfo{person}{Anubhavnidhi Abhashkumar}, \bibinfo{person}{Aaron Gember-Jacobson}, {and} \bibinfo{person}{Aditya Akella}.} \bibinfo{year}{2020}\natexlab{}.
\newblock \showarticletitle{Tiramisu: Fast multilayer network verification}. In \bibinfo{booktitle}{\emph{17th USENIX Symposium on Networked Systems Design and Implementation (NSDI 20)}}. \bibinfo{pages}{201--219}.
\newblock


\bibitem[Alsudais and Keller(2017)]%
        {alsudais2017hey}
\bibfield{author}{\bibinfo{person}{Azzam Alsudais} {and} \bibinfo{person}{Eric Keller}.} \bibinfo{year}{2017}\natexlab{}.
\newblock \showarticletitle{Hey network, can you understand me?}. In \bibinfo{booktitle}{\emph{2017 IEEE Conference on Computer Communications Workshops (INFOCOM WKSHPS)}}. IEEE, \bibinfo{pages}{193--198}.
\newblock


\bibitem[Ball et~al\mbox{.}(2014)]%
        {ball2014vericon}
\bibfield{author}{\bibinfo{person}{Thomas Ball}, \bibinfo{person}{Nikolaj Bj{\o}rner}, \bibinfo{person}{Aaron Gember}, \bibinfo{person}{Shachar Itzhaky}, \bibinfo{person}{Aleksandr Karbyshev}, \bibinfo{person}{Mooly Sagiv}, \bibinfo{person}{Michael Schapira}, {and} \bibinfo{person}{Asaf Valadarsky}.} \bibinfo{year}{2014}\natexlab{}.
\newblock \showarticletitle{Vericon: towards verifying controller programs in software-defined networks}. In \bibinfo{booktitle}{\emph{Proceedings of the 35th ACM SIGPLAN conference on programming language design and implementation}}. \bibinfo{pages}{282--293}.
\newblock


\bibitem[Barrett and Tinelli(2018)]%
        {barrett2018satisfiability}
\bibfield{author}{\bibinfo{person}{Clark Barrett} {and} \bibinfo{person}{Cesare Tinelli}.} \bibinfo{year}{2018}\natexlab{}.
\newblock \showarticletitle{Satisfiability modulo theories}.
\newblock In \bibinfo{booktitle}{\emph{Handbook of model checking}}. \bibinfo{publisher}{Springer}, \bibinfo{pages}{305--343}.
\newblock


\bibitem[Beckett et~al\mbox{.}(2017)]%
        {beckett2017general}
\bibfield{author}{\bibinfo{person}{Ryan Beckett}, \bibinfo{person}{Aarti Gupta}, \bibinfo{person}{Ratul Mahajan}, {and} \bibinfo{person}{David Walker}.} \bibinfo{year}{2017}\natexlab{}.
\newblock \showarticletitle{A general approach to network configuration verification}. In \bibinfo{booktitle}{\emph{Proceedings of the Conference of the ACM Special Interest Group on Data Communication}}. \bibinfo{pages}{155--168}.
\newblock


\bibitem[Beurer-Kellner et~al\mbox{.}(2022)]%
        {beurer2022learning}
\bibfield{author}{\bibinfo{person}{Luca Beurer-Kellner}, \bibinfo{person}{Martin Vechev}, \bibinfo{person}{Laurent Vanbever}, {and} \bibinfo{person}{Petar Veli{\v{c}}kovi{\'c}}.} \bibinfo{year}{2022}\natexlab{}.
\newblock \showarticletitle{Learning to configure computer networks with neural algorithmic reasoning}.
\newblock \bibinfo{journal}{\emph{Advances in Neural Information Processing Systems}}  \bibinfo{volume}{35} (\bibinfo{year}{2022}), \bibinfo{pages}{730--742}.
\newblock


\bibitem[Birkner et~al\mbox{.}(2018)]%
        {birkner2018net2text}
\bibfield{author}{\bibinfo{person}{R{\"u}diger Birkner}, \bibinfo{person}{Dana Drachsler-Cohen}, \bibinfo{person}{Laurent Vanbever}, {and} \bibinfo{person}{Martin Vechev}.} \bibinfo{year}{2018}\natexlab{}.
\newblock \showarticletitle{$\{$Net2Text$\}$:$\{$Query-Guided$\}$ summarization of network forwarding behaviors}. In \bibinfo{booktitle}{\emph{15th USENIX Symposium on Networked Systems Design and Implementation (NSDI 18)}}. \bibinfo{pages}{609--623}.
\newblock


\bibitem[Birkner et~al\mbox{.}(2020)]%
        {birkner2020config2spec}
\bibfield{author}{\bibinfo{person}{R{\"u}diger Birkner}, \bibinfo{person}{Dana Drachsler-Cohen}, \bibinfo{person}{Laurent Vanbever}, {and} \bibinfo{person}{Martin Vechev}.} \bibinfo{year}{2020}\natexlab{}.
\newblock \showarticletitle{$\{$Config2Spec$\}$: Mining network specifications from network configurations}. In \bibinfo{booktitle}{\emph{17th USENIX Symposium on Networked Systems Design and Implementation (NSDI 20)}}. \bibinfo{pages}{969--984}.
\newblock


\bibitem[Ceri et~al\mbox{.}(1989)]%
        {ceri1989you}
\bibfield{author}{\bibinfo{person}{Stefano Ceri}, \bibinfo{person}{Georg Gottlob}, \bibinfo{person}{Letizia Tanca}, {et~al\mbox{.}}} \bibinfo{year}{1989}\natexlab{}.
\newblock \showarticletitle{What you always wanted to know about Datalog(and never dared to ask)}.
\newblock \bibinfo{journal}{\emph{IEEE transactions on knowledge and data engineering}} \bibinfo{volume}{1}, \bibinfo{number}{1} (\bibinfo{year}{1989}), \bibinfo{pages}{146--166}.
\newblock


\bibitem[Ding et~al\mbox{.}(2024)]%
        {ding2024poster}
\bibfield{author}{\bibinfo{person}{Wenlong Ding}, \bibinfo{person}{Jianqiang Li}, \bibinfo{person}{Zhixiong Niu}, \bibinfo{person}{Huangxun Chen}, {and} \bibinfo{person}{Hong Xu}.} \bibinfo{year}{2024}\natexlab{}.
\newblock \showarticletitle{Poster: Automating Network Configuration with Natural Language Intents}. In \bibinfo{booktitle}{\emph{Proceedings of the ACM SIGCOMM 2024 Conference: Posters and Demos}}. \bibinfo{pages}{19--21}.
\newblock


\bibitem[Dong et~al\mbox{.}(2024)]%
        {dong2024abilities}
\bibfield{author}{\bibinfo{person}{Guanting Dong}, \bibinfo{person}{Hongyi Yuan}, \bibinfo{person}{Keming Lu}, \bibinfo{person}{Chengpeng Li}, \bibinfo{person}{Mingfeng Xue}, \bibinfo{person}{Dayiheng Liu}, \bibinfo{person}{Wei Wang}, \bibinfo{person}{Zheng Yuan}, \bibinfo{person}{Chang Zhou}, {and} \bibinfo{person}{Jingren Zhou}.} \bibinfo{year}{2024}\natexlab{}.
\newblock \showarticletitle{How Abilities in Large Language Models are Affected by Supervised Fine-tuning Data Composition}. In \bibinfo{booktitle}{\emph{Proceedings of the 62nd Annual Meeting of the Association for Computational Linguistics (Volume 1: Long Papers)}}. \bibinfo{pages}{177--198}.
\newblock


\bibitem[El-Hassany et~al\mbox{.}(2018)]%
        {el2018netcomplete}
\bibfield{author}{\bibinfo{person}{Ahmed El-Hassany}, \bibinfo{person}{Petar Tsankov}, \bibinfo{person}{Laurent Vanbever}, {and} \bibinfo{person}{Martin Vechev}.} \bibinfo{year}{2018}\natexlab{}.
\newblock \showarticletitle{$\{$NetComplete$\}$: Practical $\{$Network-Wide$\}$ configuration synthesis with autocompletion}. In \bibinfo{booktitle}{\emph{15th USENIX Symposium on Networked Systems Design and Implementation (NSDI 18)}}. \bibinfo{pages}{579--594}.
\newblock


\bibitem[Fang et~al\mbox{.}(2024)]%
        {fang2024large}
\bibfield{author}{\bibinfo{person}{Honglin Fang}, \bibinfo{person}{Di Zhang}, \bibinfo{person}{Can Tan}, \bibinfo{person}{Peng Yu}, \bibinfo{person}{Ying Wang}, {and} \bibinfo{person}{Wenjing Li}.} \bibinfo{year}{2024}\natexlab{}.
\newblock \showarticletitle{Large {Language} {Model} {Enhanced} {Autonomous} {Agents} for {Proactive} {Fault}-{Tolerant} {Edge} {Networks}}. In \bibinfo{booktitle}{\emph{INFOCOM}}. IEEE, \bibinfo{pages}{1--2}.
\newblock


\bibitem[Fogel et~al\mbox{.}(2015)]%
        {fogel2015general}
\bibfield{author}{\bibinfo{person}{Ari Fogel}, \bibinfo{person}{Stanley Fung}, \bibinfo{person}{Luis Pedrosa}, \bibinfo{person}{Meg Walraed-Sullivan}, \bibinfo{person}{Ramesh Govindan}, \bibinfo{person}{Ratul Mahajan}, {and} \bibinfo{person}{Todd Millstein}.} \bibinfo{year}{2015}\natexlab{}.
\newblock \showarticletitle{A general approach to network configuration analysis}. In \bibinfo{booktitle}{\emph{12th USENIX Symposium on Networked Systems Design and Implementation (NSDI 15)}}. \bibinfo{pages}{469--483}.
\newblock


\bibitem[Gember-Jacobson et~al\mbox{.}(2016)]%
        {gember2016fast}
\bibfield{author}{\bibinfo{person}{Aaron Gember-Jacobson}, \bibinfo{person}{Raajay Viswanathan}, \bibinfo{person}{Aditya Akella}, {and} \bibinfo{person}{Ratul Mahajan}.} \bibinfo{year}{2016}\natexlab{}.
\newblock \showarticletitle{Fast control plane analysis using an abstract representation}. In \bibinfo{booktitle}{\emph{Proceedings of the 2016 ACM SIGCOMM Conference}}. \bibinfo{pages}{300--313}.
\newblock


\bibitem[Gong et~al\mbox{.}(2024)]%
        {gong2024ast}
\bibfield{author}{\bibinfo{person}{Linyuan Gong}, \bibinfo{person}{Mostafa Elhoushi}, {and} \bibinfo{person}{Alvin Cheung}.} \bibinfo{year}{2024}\natexlab{}.
\newblock \showarticletitle{AST-T5: structure-aware pretraining for code generation and understanding}. In \bibinfo{booktitle}{\emph{Proceedings of the 41st International Conference on Machine Learning}}. \bibinfo{pages}{15839--15853}.
\newblock


\bibitem[Hagberg et~al\mbox{.}(2008)]%
        {hagberg2008exploring}
\bibfield{author}{\bibinfo{person}{Aric Hagberg}, \bibinfo{person}{Pieter~J Swart}, {and} \bibinfo{person}{Daniel~A Schult}.} \bibinfo{year}{2008}\natexlab{}.
\newblock \bibinfo{booktitle}{\emph{Exploring network structure, dynamics, and function using NetworkX}}.
\newblock \bibinfo{type}{{T}echnical {R}eport}. \bibinfo{institution}{Los Alamos National Laboratory (LANL), Los Alamos, NM (United States)}.
\newblock


\bibitem[Han et~al\mbox{.}(2024)]%
        {han2024netren}
\bibfield{author}{\bibinfo{person}{Rongxin Han}, \bibinfo{person}{Jingyu Wang}, \bibinfo{person}{Qi Qi}, \bibinfo{person}{Haifeng Sun}, \bibinfo{person}{Chaowei Xu}, \bibinfo{person}{Zhaoyang Wan}, \bibinfo{person}{Zirui Zhuang}, \bibinfo{person}{Yichuan Yu}, {and} \bibinfo{person}{Jianxin Liao}.} \bibinfo{year}{2024}\natexlab{}.
\newblock \showarticletitle{NetRen: Service Migration-Driven Network Renascence with Synthesizing Updated Configuration}. In \bibinfo{booktitle}{\emph{Proceedings of the 29th ACM International Conference on Architectural Support for Programming Languages and Operating Systems, Volume 3}}. \bibinfo{pages}{708--721}.
\newblock


\bibitem[Han et~al\mbox{.}(2025)]%
        {han2025network}
\bibfield{author}{\bibinfo{person}{Rongxin Han}, \bibinfo{person}{Jingyu Wang}, \bibinfo{person}{Haifeng Sun}, \bibinfo{person}{Zengteng Jiang}, \bibinfo{person}{Qi Qi}, \bibinfo{person}{Zirui Zhuang}, \bibinfo{person}{Yuan Zhang}, {and} \bibinfo{person}{Jianxin Liao}.} \bibinfo{year}{2025}\natexlab{}.
\newblock \showarticletitle{Network CoPilot: Intent-Driven Network Configuration Updating for Service Guarantee}. In \bibinfo{booktitle}{\emph{IEEE INFOCOM 2025-IEEE Conference on Computer Communications}}. IEEE, \bibinfo{pages}{1--10}.
\newblock


\bibitem[Hu et~al\mbox{.}(2022)]%
        {hu2022lora}
\bibfield{author}{\bibinfo{person}{Edward~J Hu}, \bibinfo{person}{Yelong Shen}, \bibinfo{person}{Phillip Wallis}, \bibinfo{person}{Zeyuan Allen-Zhu}, \bibinfo{person}{Yuanzhi Li}, \bibinfo{person}{Shean Wang}, \bibinfo{person}{Lu Wang}, \bibinfo{person}{Weizhu Chen}, {et~al\mbox{.}}} \bibinfo{year}{2022}\natexlab{}.
\newblock \showarticletitle{Lora: Low-rank adaptation of large language models.}
\newblock \bibinfo{journal}{\emph{ICLR}} \bibinfo{volume}{1}, \bibinfo{number}{2} (\bibinfo{year}{2022}), \bibinfo{pages}{3}.
\newblock


\bibitem[Jacobs et~al\mbox{.}(2021)]%
        {jacobs2021hey}
\bibfield{author}{\bibinfo{person}{Arthur~S Jacobs}, \bibinfo{person}{Ricardo~J Pfitscher}, \bibinfo{person}{Rafael~H Ribeiro}, \bibinfo{person}{Ronaldo~A Ferreira}, \bibinfo{person}{Lisandro~Z Granville}, \bibinfo{person}{Walter Willinger}, {and} \bibinfo{person}{Sanjay~G Rao}.} \bibinfo{year}{2021}\natexlab{}.
\newblock \showarticletitle{Hey, lumi! using natural language for $\{$intent-based$\}$ network management}. In \bibinfo{booktitle}{\emph{2021 usenix annual technical conference (usenix atc 21)}}. \bibinfo{pages}{625--639}.
\newblock


\bibitem[Jelaca et~al\mbox{.}(2025)]%
        {jelaca2025automated}
\bibfield{author}{\bibinfo{person}{Aleksa Jelaca}, \bibinfo{person}{Ying Jiao}, \bibinfo{person}{Chang Tian}, {and} \bibinfo{person}{Marie-Francine Moens}.} \bibinfo{year}{2025}\natexlab{}.
\newblock \showarticletitle{Automated prompt generation for creative and counterfactual text-to-image synthesis}.
\newblock \bibinfo{journal}{\emph{arXiv preprint arXiv:2509.21375}} (\bibinfo{year}{2025}).
\newblock


\bibitem[Jiang et~al\mbox{.}(2024)]%
        {jiang2024survey}
\bibfield{author}{\bibinfo{person}{Juyong Jiang}, \bibinfo{person}{Fan Wang}, \bibinfo{person}{Jiasi Shen}, \bibinfo{person}{Sungju Kim}, {and} \bibinfo{person}{Sunghun Kim}.} \bibinfo{year}{2024}\natexlab{}.
\newblock \showarticletitle{A survey on large language models for code generation}.
\newblock \bibinfo{journal}{\emph{arXiv preprint arXiv:2406.00515}} (\bibinfo{year}{2024}).
\newblock


\bibitem[Kheradmand(2020)]%
        {kheradmand2020automatic}
\bibfield{author}{\bibinfo{person}{Ali Kheradmand}.} \bibinfo{year}{2020}\natexlab{}.
\newblock \showarticletitle{Automatic inference of high-level network intents by mining forwarding patterns}. In \bibinfo{booktitle}{\emph{Proceedings of the Symposium on SDN Research}}. \bibinfo{pages}{27--33}.
\newblock


\bibitem[Knight et~al\mbox{.}(2011)]%
        {6027859}
\bibfield{author}{\bibinfo{person}{S. Knight}, \bibinfo{person}{H.X. Nguyen}, \bibinfo{person}{N. Falkner}, \bibinfo{person}{R. Bowden}, {and} \bibinfo{person}{M. Roughan}.} \bibinfo{year}{2011}\natexlab{}.
\newblock \showarticletitle{The Internet Topology Zoo}.
\newblock \bibinfo{journal}{\emph{Selected Areas in Communications, IEEE Journal on}} \bibinfo{volume}{29}, \bibinfo{number}{9} (\bibinfo{date}{october} \bibinfo{year}{2011}), \bibinfo{pages}{1765 --1775}.
\newblock
\showISSN{0733-8716}
\href{https://doi.org/10.1109/JSAC.2011.111002}{doi:\nolinkurl{10.1109/JSAC.2011.111002}}


\bibitem[Kou et~al\mbox{.}(2025)]%
        {kou2025gia}
\bibfield{author}{\bibinfo{person}{Shiwen Kou}, \bibinfo{person}{Chungang Yang}, {and} \bibinfo{person}{Mohan Gurusamy}.} \bibinfo{year}{2025}\natexlab{}.
\newblock \showarticletitle{GIA: LLM-{Enabled} {Generative} {Intent} {Abstraction} to {Enhance} {Adaptability} for {Intent}-{Driven} {Networks}}.
\newblock \bibinfo{journal}{\emph{IEEE Transactions on Cognitive Communications and Networking}}  \bibinfo{volume}{11} (\bibinfo{year}{2025}), \bibinfo{pages}{999--1012}.
\newblock


\bibitem[Laurent and Rivest(1976)]%
        {laurent1976constructing}
\bibfield{author}{\bibinfo{person}{Hyafil Laurent} {and} \bibinfo{person}{Ronald~L Rivest}.} \bibinfo{year}{1976}\natexlab{}.
\newblock \showarticletitle{Constructing optimal binary decision trees is NP-complete}.
\newblock \bibinfo{journal}{\emph{Information processing letters}} \bibinfo{volume}{5}, \bibinfo{number}{1} (\bibinfo{year}{1976}), \bibinfo{pages}{15--17}.
\newblock


\bibitem[Lewis et~al\mbox{.}(2020)]%
        {lewis2020retrieval}
\bibfield{author}{\bibinfo{person}{Patrick Lewis}, \bibinfo{person}{Ethan Perez}, \bibinfo{person}{Aleksandra Piktus}, \bibinfo{person}{Fabio Petroni}, \bibinfo{person}{Vladimir Karpukhin}, \bibinfo{person}{Naman Goyal}, \bibinfo{person}{Heinrich K{\"u}ttler}, \bibinfo{person}{Mike Lewis}, \bibinfo{person}{Wen-tau Yih}, \bibinfo{person}{Tim Rockt{\"a}schel}, {et~al\mbox{.}}} \bibinfo{year}{2020}\natexlab{}.
\newblock \showarticletitle{Retrieval-augmented generation for knowledge-intensive nlp tasks}.
\newblock \bibinfo{journal}{\emph{Advances in neural information processing systems}}  \bibinfo{volume}{33} (\bibinfo{year}{2020}), \bibinfo{pages}{9459--9474}.
\newblock


\bibitem[Li et~al\mbox{.}(2021)]%
        {li2021paint4poem}
\bibfield{author}{\bibinfo{person}{Dan Li}, \bibinfo{person}{Shuai Wang}, \bibinfo{person}{Jie Zou}, \bibinfo{person}{Chang Tian}, \bibinfo{person}{Elisha Nieuwburg}, \bibinfo{person}{Fengyuan Sun}, {and} \bibinfo{person}{Evangelos Kanoulas}.} \bibinfo{year}{2021}\natexlab{}.
\newblock \showarticletitle{Paint4Poem: A dataset for artistic visualization of classical Chinese poems}.
\newblock \bibinfo{journal}{\emph{arXiv preprint arXiv:2109.11682}} (\bibinfo{year}{2021}).
\newblock


\bibitem[Liu et~al\mbox{.}(2025)]%
        {liu2025cegs}
\bibfield{author}{\bibinfo{person}{Jianmin Liu}, \bibinfo{person}{Li Chen}, \bibinfo{person}{Dan Li}, {and} \bibinfo{person}{Yukai Miao}.} \bibinfo{year}{2025}\natexlab{}.
\newblock \showarticletitle{$\{$CEGS$\}$: Configuration Example Generalizing Synthesizer}. In \bibinfo{booktitle}{\emph{22nd USENIX Symposium on Networked Systems Design and Implementation (NSDI 25)}}. \bibinfo{pages}{1327--1347}.
\newblock


\bibitem[Magzhan and Jani(2013)]%
        {magzhan2013review}
\bibfield{author}{\bibinfo{person}{Kairanbay Magzhan} {and} \bibinfo{person}{Hajar~Mat Jani}.} \bibinfo{year}{2013}\natexlab{}.
\newblock \showarticletitle{A review and evaluations of shortest path algorithms}.
\newblock \bibinfo{journal}{\emph{Int. J. Sci. Technol. Res}} \bibinfo{volume}{2}, \bibinfo{number}{6} (\bibinfo{year}{2013}), \bibinfo{pages}{99--104}.
\newblock


\bibitem[Maier et~al\mbox{.}(2018)]%
        {maier2018datalog}
\bibfield{author}{\bibinfo{person}{David Maier}, \bibinfo{person}{K~Tuncay Tekle}, \bibinfo{person}{Michael Kifer}, {and} \bibinfo{person}{David~S Warren}.} \bibinfo{year}{2018}\natexlab{}.
\newblock \showarticletitle{Datalog: concepts, history, and outlook}.
\newblock In \bibinfo{booktitle}{\emph{Declarative Logic Programming: Theory, Systems, and Applications}}. \bibinfo{pages}{3--100}.
\newblock


\bibitem[Martinez et~al\mbox{.}(2014)]%
        {martinez2014network}
\bibfield{author}{\bibinfo{person}{Anny Martinez}, \bibinfo{person}{Marcelo Yannuzzi}, \bibinfo{person}{V{\'\i}ctor L{\'o}pez}, \bibinfo{person}{Diego L{\'o}pez}, \bibinfo{person}{Wilson Ram{\'\i}rez}, \bibinfo{person}{Ren{\'e} Serral-Graci{\`a}}, \bibinfo{person}{Xavi Masip-Bruin}, \bibinfo{person}{Maciej Maciejewski}, {and} \bibinfo{person}{J{\"o}rn Altmann}.} \bibinfo{year}{2014}\natexlab{}.
\newblock \showarticletitle{Network management challenges and trends in multi-layer and multi-vendor settings for carrier-grade networks}.
\newblock \bibinfo{journal}{\emph{IEEE Communications Surveys \& Tutorials}} \bibinfo{volume}{16}, \bibinfo{number}{4} (\bibinfo{year}{2014}), \bibinfo{pages}{2207--2230}.
\newblock


\bibitem[Martinez(2012)]%
        {martinez2012part}
\bibfield{author}{\bibinfo{person}{Angel~R Martinez}.} \bibinfo{year}{2012}\natexlab{}.
\newblock \showarticletitle{Part-of-speech tagging}.
\newblock \bibinfo{journal}{\emph{Wiley Interdisciplinary Reviews: Computational Statistics}} \bibinfo{volume}{4}, \bibinfo{number}{1} (\bibinfo{year}{2012}), \bibinfo{pages}{107--113}.
\newblock


\bibitem[Mekrache et~al\mbox{.}(2024)]%
        {mekrache2024intentbased}
\bibfield{author}{\bibinfo{person}{Abdelkader Mekrache}, \bibinfo{person}{Adlen Ksentini}, {and} \bibinfo{person}{Christos Verikoukis}.} \bibinfo{year}{2024}\natexlab{}.
\newblock \showarticletitle{Intent-{Based} {Management} of {Next}-{Generation} {Networks}: an {LLM}-centric {Approach}}.
\newblock \bibinfo{journal}{\emph{IEEE Network}} (\bibinfo{year}{2024}), \bibinfo{pages}{1--1}.
\newblock


\bibitem[Mondal et~al\mbox{.}(2023a)]%
        {mondal2023llms}
\bibfield{author}{\bibinfo{person}{Rajdeep Mondal}, \bibinfo{person}{Alan Tang}, \bibinfo{person}{Ryan Beckett}, \bibinfo{person}{Todd Millstein}, {and} \bibinfo{person}{George Varghese}.} \bibinfo{year}{2023}\natexlab{a}.
\newblock \showarticletitle{What do LLMs need to synthesize correct router configurations?}. In \bibinfo{booktitle}{\emph{Proceedings of the 22nd ACM Workshop on Hot Topics in Networks}}. \bibinfo{pages}{189--195}.
\newblock


\bibitem[Mondal et~al\mbox{.}(2023b)]%
        {mondal2023what}
\bibfield{author}{\bibinfo{person}{Rajdeep Mondal}, \bibinfo{person}{Alan Tang}, \bibinfo{person}{Ryan Beckett}, \bibinfo{person}{Todd Millstein}, {and} \bibinfo{person}{George Varghese}.} \bibinfo{year}{2023}\natexlab{b}.
\newblock \showarticletitle{What do {LLMs} need to {Synthesize} {Correct} {Router} {Configurations}?}. In \bibinfo{booktitle}{\emph{HotNets}}. ACM.
\newblock


\bibitem[Moy(1998)]%
        {moy1998ospf}
\bibfield{author}{\bibinfo{person}{John Moy}.} \bibinfo{year}{1998}\natexlab{}.
\newblock \showarticletitle{OSPF version 2, April 1998}.
\newblock \bibinfo{journal}{\emph{Internet Engineering Taskforce RFC}}  \bibinfo{volume}{2328} (\bibinfo{year}{1998}).
\newblock


\bibitem[Reimers et~al\mbox{.}(1908)]%
        {reimers1908sentence}
\bibfield{author}{\bibinfo{person}{Nils Reimers}, \bibinfo{person}{I~Sentence-BERT Gurevych}, {et~al\mbox{.}}} \bibinfo{year}{1908}\natexlab{}.
\newblock \showarticletitle{Sentence embeddings using siamese BERT-networks. arXiv 2019}.
\newblock \bibinfo{journal}{\emph{arXiv preprint arXiv:1908.10084}}  \bibinfo{volume}{10} (\bibinfo{year}{1908}).
\newblock


\bibitem[Rekhter et~al\mbox{.}(2006)]%
        {rekhter2006rfc}
\bibfield{author}{\bibinfo{person}{Yakov Rekhter}, \bibinfo{person}{Tony Li}, {and} \bibinfo{person}{Susan Hares}.} \bibinfo{year}{2006}\natexlab{}.
\newblock \bibinfo{title}{RFC 4271: A border gateway protocol 4 (BGP-4)}.
\newblock


\bibitem[Schildermans et~al\mbox{.}(2025)]%
        {schildermans2025structured}
\bibfield{author}{\bibinfo{person}{Sander Schildermans}, \bibinfo{person}{Chang Tian}, \bibinfo{person}{Ying Jiao}, {and} \bibinfo{person}{Marie-Francine Moens}.} \bibinfo{year}{2025}\natexlab{}.
\newblock \showarticletitle{Structured information for improving spatial relationships in text-to-image generation}.
\newblock \bibinfo{journal}{\emph{arXiv preprint arXiv:2509.15962}} (\bibinfo{year}{2025}).
\newblock


\bibitem[Tang et~al\mbox{.}(2023)]%
        {tang2023lightyear}
\bibfield{author}{\bibinfo{person}{Alan Tang}, \bibinfo{person}{Ryan Beckett}, \bibinfo{person}{Steven Benaloh}, \bibinfo{person}{Karthick Jayaraman}, \bibinfo{person}{Tejas Patil}, \bibinfo{person}{Todd Millstein}, {and} \bibinfo{person}{George Varghese}.} \bibinfo{year}{2023}\natexlab{}.
\newblock \showarticletitle{Lightyear: Using modularity to scale bgp control plane verification}. In \bibinfo{booktitle}{\emph{Proceedings of the ACM SIGCOMM 2023 Conference}}. \bibinfo{pages}{94--107}.
\newblock


\bibitem[Tian et~al\mbox{.}(2025a)]%
        {tian2025large}
\bibfield{author}{\bibinfo{person}{Chang Tian}, \bibinfo{person}{Matthew~B Blaschko}, \bibinfo{person}{Mingzhe Xing}, \bibinfo{person}{Xiuxing Li}, \bibinfo{person}{Yinliang Yue}, {and} \bibinfo{person}{Marie-Francine Moens}.} \bibinfo{year}{2025}\natexlab{a}.
\newblock \showarticletitle{Large language models reasoning abilities under non-ideal conditions after rl-fine-tuning}.
\newblock \bibinfo{journal}{\emph{arXiv preprint arXiv:2508.04848}} (\bibinfo{year}{2025}).
\newblock


\bibitem[Tian et~al\mbox{.}(2024a)]%
        {tian2024generic}
\bibfield{author}{\bibinfo{person}{Chang Tian}, \bibinfo{person}{Matthew~B Blaschko}, \bibinfo{person}{Wenpeng Yin}, \bibinfo{person}{Mingzhe Xing}, \bibinfo{person}{Yinliang Yue}, {and} \bibinfo{person}{Marie-Francine Moens}.} \bibinfo{year}{2024}\natexlab{a}.
\newblock \showarticletitle{A generic method for fine-grained category discovery in natural language texts}.
\newblock \bibinfo{journal}{\emph{arXiv preprint arXiv:2406.13103}} (\bibinfo{year}{2024}).
\newblock


\bibitem[Tian et~al\mbox{.}(2025b)]%
        {tian2025using}
\bibfield{author}{\bibinfo{person}{Chang Tian}, \bibinfo{person}{Mingzhe Xing}, \bibinfo{person}{Zenglin Shi}, \bibinfo{person}{Matthew~B Blaschko}, \bibinfo{person}{Yinliang Yue}, {and} \bibinfo{person}{Marie-Francine Moens}.} \bibinfo{year}{2025}\natexlab{b}.
\newblock \showarticletitle{Using Causality for Enhanced Prediction of Web Traffic Time Series}.
\newblock \bibinfo{journal}{\emph{arXiv preprint arXiv:2502.00612}} (\bibinfo{year}{2025}).
\newblock


\bibitem[Tian et~al\mbox{.}(2024b)]%
        {tian2024fighting}
\bibfield{author}{\bibinfo{person}{Chang Tian}, \bibinfo{person}{Wenpeng Yin}, \bibinfo{person}{Dan Li}, {and} \bibinfo{person}{Marie-Francine Moens}.} \bibinfo{year}{2024}\natexlab{b}.
\newblock \showarticletitle{Fighting against the repetitive training and sample dependency problem in few-shot named entity recognition}.
\newblock \bibinfo{journal}{\emph{Ieee Access}}  \bibinfo{volume}{12} (\bibinfo{year}{2024}), \bibinfo{pages}{37600--37614}.
\newblock


\bibitem[Tian et~al\mbox{.}(2022)]%
        {tian2022anti}
\bibfield{author}{\bibinfo{person}{Chang Tian}, \bibinfo{person}{Wenpeng Yin}, {and} \bibinfo{person}{Marie-Francine Moens}.} \bibinfo{year}{2022}\natexlab{}.
\newblock \showarticletitle{Anti-overestimation dialogue policy learning for task-completion dialogue system}.
\newblock \bibinfo{journal}{\emph{arXiv preprint arXiv:2207.11762}} (\bibinfo{year}{2022}).
\newblock


\bibitem[Touvron et~al\mbox{.}(2023)]%
        {touvron2023llama}
\bibfield{author}{\bibinfo{person}{Hugo Touvron}, \bibinfo{person}{Thibaut Lavril}, \bibinfo{person}{Gautier Izacard}, \bibinfo{person}{Xavier Martinet}, \bibinfo{person}{Marie-Anne Lachaux}, \bibinfo{person}{Timoth{\'e}e Lacroix}, \bibinfo{person}{Baptiste Rozi{\`e}re}, \bibinfo{person}{Naman Goyal}, \bibinfo{person}{Eric Hambro}, \bibinfo{person}{Faisal Azhar}, {et~al\mbox{.}}} \bibinfo{year}{2023}\natexlab{}.
\newblock \showarticletitle{Llama: Open and efficient foundation language models}.
\newblock \bibinfo{journal}{\emph{arXiv preprint arXiv:2302.13971}} (\bibinfo{year}{2023}).
\newblock


\bibitem[Wang et~al\mbox{.}(2024)]%
        {wang2024netconfeval}
\bibfield{author}{\bibinfo{person}{Changjie Wang}, \bibinfo{person}{Mariano Scazzariello}, \bibinfo{person}{Alireza Farshin}, \bibinfo{person}{Simone Ferlin}, \bibinfo{person}{Dejan Kosti{\' c}}, {and} \bibinfo{person}{Marco Chiesa}.} \bibinfo{year}{2024}\natexlab{}.
\newblock \showarticletitle{NetConfEval: Can {LLMs} {Facilitate} {Network} {Configuration}?}
\newblock \bibinfo{journal}{\emph{CoNEXT}} \bibinfo{volume}{2}, \bibinfo{number}{CoNEXT2} (\bibinfo{year}{2024}), \bibinfo{pages}{1--25}.
\newblock


\bibitem[Wang et~al\mbox{.}(2025)]%
        {wang2025document}
\bibfield{author}{\bibinfo{person}{Zhitong Wang}, \bibinfo{person}{Cheng Gao}, \bibinfo{person}{Chaojun Xiao}, \bibinfo{person}{Yufei Huang}, \bibinfo{person}{Shuzheng Si}, \bibinfo{person}{Kangyang Luo}, \bibinfo{person}{Yuzhuo Bai}, \bibinfo{person}{Wenhao Li}, \bibinfo{person}{Tangjian Duan}, \bibinfo{person}{Chuancheng Lv}, {et~al\mbox{.}}} \bibinfo{year}{2025}\natexlab{}.
\newblock \showarticletitle{Document Segmentation Matters for Retrieval-Augmented Generation}. In \bibinfo{booktitle}{\emph{Findings of the Association for Computational Linguistics: ACL 2025}}. \bibinfo{pages}{8063--8075}.
\newblock


\bibitem[Wei et~al\mbox{.}(2025)]%
        {wei2025intaintentbasedtranslationnetwork}
\bibfield{author}{\bibinfo{person}{Yunze Wei}, \bibinfo{person}{Xiaohui Xie}, \bibinfo{person}{Tianshuo Hu}, \bibinfo{person}{Yiwei Zuo}, \bibinfo{person}{Xinyi Chen}, \bibinfo{person}{Kaiwen Chi}, {and} \bibinfo{person}{Yong Cui}.} \bibinfo{year}{2025}\natexlab{}.
\newblock \bibinfo{title}{INTA: Intent-Based Translation for Network Configuration with LLM Agents}.
\newblock
\showeprint[arxiv]{2501.08760}~[cs.NI]
\urldef\tempurl%
\url{https://arxiv.org/abs/2501.08760}
\showURL{%
\tempurl}


\bibitem[Wu et~al\mbox{.}(2024)]%
        {wu2024netllm}
\bibfield{author}{\bibinfo{person}{Duo Wu}, \bibinfo{person}{Xianda Wang}, \bibinfo{person}{Yaqi Qiao}, \bibinfo{person}{Zhi Wang}, \bibinfo{person}{Junchen Jiang}, \bibinfo{person}{Shuguang Cui}, {and} \bibinfo{person}{Fangxin Wang}.} \bibinfo{year}{2024}\natexlab{}.
\newblock \showarticletitle{NetLLM: Adapting {Large} {Language} {Models} for {Networking}}. In \bibinfo{booktitle}{\emph{SIGCOMM}}, Vol.~\bibinfo{volume}{33}. ACM, \bibinfo{pages}{661--678}.
\newblock


\bibitem[Wu et~al\mbox{.}(2020)]%
        {wu2020comprehensive}
\bibfield{author}{\bibinfo{person}{Zonghan Wu}, \bibinfo{person}{Shirui Pan}, \bibinfo{person}{Fengwen Chen}, \bibinfo{person}{Guodong Long}, \bibinfo{person}{Chengqi Zhang}, {and} \bibinfo{person}{Philip~S Yu}.} \bibinfo{year}{2020}\natexlab{}.
\newblock \showarticletitle{A comprehensive survey on graph neural networks}.
\newblock \bibinfo{journal}{\emph{IEEE transactions on neural networks and learning systems}} \bibinfo{volume}{32}, \bibinfo{number}{1} (\bibinfo{year}{2020}), \bibinfo{pages}{4--24}.
\newblock


\bibitem[Xing et~al\mbox{.}(2024)]%
        {xing2024understanding}
\bibfield{author}{\bibinfo{person}{Mingzhe Xing}, \bibinfo{person}{Rongkai Zhang}, \bibinfo{person}{Hui Xue}, \bibinfo{person}{Qi Chen}, \bibinfo{person}{Fan Yang}, {and} \bibinfo{person}{Zhen Xiao}.} \bibinfo{year}{2024}\natexlab{}.
\newblock \showarticletitle{Understanding the weakness of large language model agents within a complex android environment}. In \bibinfo{booktitle}{\emph{Proceedings of the 30th ACM SIGKDD Conference on Knowledge Discovery and Data Mining}}. \bibinfo{pages}{6061--6072}.
\newblock


\bibitem[Yang et~al\mbox{.}(2025)]%
        {yang2025qwen3}
\bibfield{author}{\bibinfo{person}{An Yang}, \bibinfo{person}{Anfeng Li}, \bibinfo{person}{Baosong Yang}, \bibinfo{person}{Beichen Zhang}, \bibinfo{person}{Binyuan Hui}, \bibinfo{person}{Bo Zheng}, \bibinfo{person}{Bowen Yu}, \bibinfo{person}{Chang Gao}, \bibinfo{person}{Chengen Huang}, \bibinfo{person}{Chenxu Lv}, {et~al\mbox{.}}} \bibinfo{year}{2025}\natexlab{}.
\newblock \showarticletitle{Qwen3 technical report}.
\newblock \bibinfo{journal}{\emph{arXiv preprint arXiv:2505.09388}} (\bibinfo{year}{2025}).
\newblock


\bibitem[Zhao et~al\mbox{.}(2023)]%
        {zhao2023survey}
\bibfield{author}{\bibinfo{person}{Wayne~Xin Zhao}, \bibinfo{person}{Kun Zhou}, \bibinfo{person}{Junyi Li}, \bibinfo{person}{Tianyi Tang}, \bibinfo{person}{Xiaolei Wang}, \bibinfo{person}{Yupeng Hou}, \bibinfo{person}{Yingqian Min}, \bibinfo{person}{Beichen Zhang}, \bibinfo{person}{Junjie Zhang}, \bibinfo{person}{Zican Dong}, {et~al\mbox{.}}} \bibinfo{year}{2023}\natexlab{}.
\newblock \showarticletitle{A survey of large language models}.
\newblock \bibinfo{journal}{\emph{arXiv preprint arXiv:2303.18223}} \bibinfo{volume}{1}, \bibinfo{number}{2} (\bibinfo{year}{2023}).
\newblock


\bibitem[Zhou et~al\mbox{.}(2025)]%
        {zhou2025controllable}
\bibfield{author}{\bibinfo{person}{Fan Zhou}, \bibinfo{person}{Chang Tian}, {and} \bibinfo{person}{Tim Van~de Cruys}.} \bibinfo{year}{2025}\natexlab{}.
\newblock \showarticletitle{Controllable Stylistic Text Generation with Train-Time Attribute-Regularized Diffusion}.
\newblock \bibinfo{journal}{\emph{arXiv preprint arXiv:2510.06386}} (\bibinfo{year}{2025}).
\newblock


\end{thebibliography}
